\documentclass[twoside,leqno,twocolumn]{article}
\pdfoutput=1
\usepackage{ltexpprt}
\usepackage[numbers]{natbib}
\usepackage[leqno]{amsmath}
\usepackage{amssymb}
\newcommand{\reals}{\mathbb{R}}
\usepackage{mathtools}
\mathtoolsset{mathic}
\usepackage{subfig}
\usepackage{url}
\usepackage[inline]{enumitem} % inline lists
\usepackage[svgnames]{xcolor}
\usepackage[pdftex]{hyperref} 
\hypersetup{colorlinks,linkcolor=red,citecolor=purple,urlcolor=DarkBlue}
\usepackage[final,inline,nomargin,index]{fixme}
\fxsetup{theme=color,mode=multiuser,inlineface=\itshape,envface=\itshape}
\FXRegisterAuthor{sv}{asv}{\colorbox{gray!10!white}{\color{black}Suresh}}
\FXRegisterAuthor{sf}{asf}{\colorbox{blue!10!white}{\color{black}Sorelle}}
\FXRegisterAuthor{cs}{acs}{\colorbox{red!10!white}{\color{black}Carlos}}
\FXRegisterAuthor{ma}{ama}{\colorbox{green!10!white}{\color{black}Mohsen}}

\DeclareMathOperator*{\argmax}{arg\,max}

\newcommand{\RNum}[1]{\uppercase\expandafter{\romannumeral #1\relax}}
\newcommand{\bignote}[1]{%
  \begin{center}
    \vspace{-1em}
    \noindent\colorbox{gray!20}{%
      \begin{minipage}{\columnwidth}%
        \noindent #1%
      \end{minipage}%
    }
  \end{center}
}

\title{Fairness in representation: quantifying stereotyping as a
  representational harm\footnote{This research was funded in part by the NSF
    under grants IIS-1633387, IIS-1633724, IIS-1513651 and IIS-1815238.}}
\author{Mohsen Abbasi\thanks{University of Utah,\url{mailto:mohsena@cs.utah.edu}} \and Sorelle A. Friedler\thanks{Haverford College, \url{mailto:sorelle@cs.haverford.edu}} \and Carlos Scheidegger\thanks{University of Arizona, \url{mailto:cscheid@cs.arizona.edu}} \and Suresh Venkatasubramanian\thanks{University of Utah, \url{mailto:suresh@cs.utah.edu}}}
\date{}

\begin{document}
\maketitle

\fancyfoot[R]{\scriptsize{Copyright \textcopyright\ 2019 by SIAM\\
Unauthorized reproduction of this article is prohibited}}

\begin{abstract} \small\baselineskip=9pt

While harms of allocation have been increasingly studied as part of the subfield of algorithmic fairness, harms of representation have received considerably less attention.  In this paper, we formalize two notions of stereotyping and show how they manifest in later allocative harms within the machine learning pipeline.  We also propose mitigation strategies and demonstrate their effectiveness on synthetic datasets.
\end{abstract}

\section{Introduction}
\label{sec:introduction}

In the rapidly growing area of fairness, accountability and transparency in machine learning, one of the fundamental questions is the problem of \emph{discrimination}: are there disparities between social groups in the way decisions are made? This question has been typically studied as a \emph{harm of allocation}: a problem in the way a learned model allocates decisions to entities. In a talk at NIPS 2017\cite{crawford}, Kate Crawford proposed studying instead harms of \emph{representation}: the ways in which individuals might be \emph{represented} differently in a feature space even before training a model.  For example, she describes the representation of Black people as inherently more criminal as a harm whether or not hiring decisions are made based on that representation.  Similarly, the work that has been done showing that word embeddings contain gender bias \cite{Bolukbasi:2016:MCP:3157382.3157584,caliskan_semantics_2017,speer} identifies a harm of representation.

To study representational harm and how to minimize it, we must quantify it. Friedler et al. \cite{friedler2016possibility} provide a framework that explicitly calls out the distinction between the \emph{construct space}, the desired representation of individuals, and the \emph{observed space}, the measured attributes. They then propose measures of distortion between these spaces as a way to measure structural bias in representation.

Harms of representation come in many different forms. Perhaps the most ubiquitous one is \emph{stereotyping} -- the tendency to assign characteristics to \emph{all} members of a group based on stereotypical features shared by a few.  In this paper, we focus on quantifying stereotyping as a form of representation distortion. Our goal is to apply the general framework of Friedler et al. in this specific context, in order to design measures that are specifically sensitive to stereotyping effects. 

\subsection{Our Contributions.}
\label{sec:our-contributions}

Our main contributions in this paper are:

\begin{itemize}[itemsep=1pt]
\item A formal mechanism for stereotyping as a function from construct to
  observed space. This mechanism can be interpreted probabilistically or
  geometrically: in its former form it aligns with literature in psychology that
  explores how people form stereotypes. 
\item A demonstration of the effects of stereotyping in model building. 
\item A proposal for mitigating the effects of stereotyping -- in effect an
  attempt to ``invert'' stereotyping as defined above -- and experimental evidence demonstrating its effectiveness.
\end{itemize}

%%% Local Variables:
%%% mode: latex
%%% TeX-master: "stereo"
%%% End:
	
\section{Literature review}
\label{sec:lit-review}

In psychology, a stereotype is defined as an over-generalized belief about a particular group or class of people \citep{cardwell1999}. Stereotypes can be positive: \emph{Asians are good at math}, or negative: \emph{African-American names are more associated with criminal backgrounds} \citep{sweeney2013discrimination}; Stereotypes are not limited to just racial groups and they can change over time \citep{katz1933verbal}.
%: \emph{Women being bad at driving} or \emph{Labeling LGBTQ literature as "adult content" by Amazon}. 
One of the central purposes served by applying stereotypes is to simplify our social world as they reduce the amount of information processing we have to do when faced with situations similar to our past experiences \citep{hilton1996stereotypes}.  Though stereotypes can be seen as helping people respond to different social situations more promptly, they make us overlook individual differences; this can lead to prejudice \citep{statt2002}. 

% Sociology literature approach mostly views stereotyping to be harmful in nature. \citep{steele2010whistling} defines \emph{identity contingencies} as "things you have to deal with in a situation because you have a given social identity, because you are old, young, gay, a white male, a woman, Latino, politically conservative or liberal, a cancer patient and so on". They argue that we're all well aware of our social identity and stereotype threat arises when an individual becomes consumed with the fear of confirming a particular stereotype by his or her actions (e.g., a female performing poorly on a math exam).

%In \citep{katz1933verbal}, the stereotypical mindset of Americans towards other races was investigated by providing a questionnaire to college students. In 1933 , Japanese were seen as \emph{industrious, intelligent, progressive, shrewd and sly}. Repeating the same study, the stereotypes attributed to the Japanese in 1951 were \emph{imitative, sly, extremely, nationalistic, treacherous}; and prominent Japanese stereotypes changed to \emph{industrious, ambitious, efficient, intelligent and progressive} in 1967. While the stereotypes themselves changed, the belief about particular ethnic groups having particular characteristics existed in all of the studies. 
The approach we take in studying stereotypes is inspired by the literature on social cognition. This approach defines stereotypes as beliefs about the characteristics, attributes, and behaviors of members of certain groups and views stereotypic thinking as a mechanism which serves a variety of cognitive and motivational processes e.g. simplifying information processing to save cognitive resources \citep{hilton1996stereotypes}. Various models have been proposed and used to represent stereotyping. In the \emph{prototype} model, people store and use an abstraction of stereotyped group's typical features and judge the the said group members by their similarity to this prototype \citep{cantor1979prototypes}.  In the \emph{exemplar} model, people use specific, real world individuals as representatives for the groups. As a group might have a number of exemplars; which one comes to mind when the stereotyping process is being activated depends on the context and situation in which the encounter with the stereotyped group member has occurred \citep{smith1992exemplar}.  In \emph{associative networks}, stereotypes are considered as linked attributes which are extensively interconnected \citep{manis1988stereotypes}. In the \emph{schemas} model, stereotypes are thought of as highly generalized beliefs about group members with no specific abstraction or exemplar for an attribute tied with such beliefs. Finally, in the \emph{representativeness} model, stereotyping is defined as distorted perception of the relative frequency of a \textit{type} in the stereotyped group compared to that of a base group \citep{bordalo2016stereotypes}. This definition is based on the \emph{representativeness heuristic} due to Kahnemann and Tversky \citep{kahneman1972subjective,kahneman1973psychology}, a similarity heuristic that people rely on to judge the likelihood of uncertain events, instead of following the principles of probability theory~\citep{tversky1974judgment}.

Stereotyping also makes an appearance in the economic literature on statistical discrimination~\citep{becker1957economics, arrow1973theory}. Statistical discrimination describes the process where employers, unable to perfectly assess worker's productivity at the time of hiring, use information such as sex and race as proxies for the expected productivity. This is estimated by the employer's prior knowledge of the average productivity of the group the worker belongs to. In other words, the stereotyping mechanism described here is the replacement of individual scores by a single aggregated score over a group, where the score is perceived as relevant for the individual. This literature views stereotyping as a rational response to insufficient information about individuals, rather than as a choice of representation that might distort outcomes.

The literature on algorithmic fairness has for the most part focused on bias caused by skew in training data distributions and the training process itself, and has quantified the bias in terms of fairness measures that are typically outcome-based (see, e.g., \cite{Calders10NaiveBayes, hardt2016equality, zafar2017mistreatment, Kamishima12Fairness} and surveys \cite{Romei13Multidisciplinary, zliobaite2015survey}). Some of the research in this area has taken a representational, preprocessing approach to reversing training data skew \cite{Feldman2015DisparateImpact, Madras2018,icml2013_zemel13,brunet2018,edwards2015} by changing the inputs so that a classifier finds fair outcomes. In the context of unsupervised learning, recent work on fair clustering\cite{chierichetti2017fair,Schmidt18,Bera19} and PCA\cite{samadi_price_2018,olfat_convex_2018} seeks to generate a modified representation of the input points so that the new representation (clustering or reduced dimension) satisfies a notion of ``balance'' with respect to groups. In all of these, the goal is to use representation to guide (fair) learning, rather than look at skew in the representation itself. 

There have been a few works that look at bias that emerges from the representation process, most notably when looking at \emph{learned} representations that come from word-vector embeddings\cite{Bolukbasi:2016:MCP:3157382.3157584,caliskan_semantics_2017,speer,de-arteaga_bias_2019}. The goal of these methods is to show how biases in language are preserved after doing such an embedding. 

%%% Local Variables:
%%% mode: latex
%%% TeX-master: "stereo"
%%% End:

\section{Modeling Stereotypes}
\label{sec:model}

We now propose mechanisms by which stereotyping might occur. We first present a novel geometric approach to stereotyping (and a variant on it), and then review a \emph{probabilistic} approach first proposed in \citep{bordalo2016stereotypes}. Finally, we show that these different perspectives on stereotyping can be unified in a common algebraic framework. Each mechanism will have an associated stereotyping measure, with larger values indicating a greater degree of stereotyping. 

\subsection{Stereotyping via Exemplars.}
Stereotyping via exemplars refers to generalizing features attributed to a small subset of a group, called \emph{stereotypic exemplars}, to all of its members. In the simplest version of stereotyping by exemplars, a single exemplar \emph{pulls} points towards itself, so that in the observed representation, points from one group are perceived to be closer to the exemplar (and thus closer in feature space) than they actually are. 

\begin{figure}[htbp]%
    \centering
	\includegraphics[width=0.8\columnwidth]{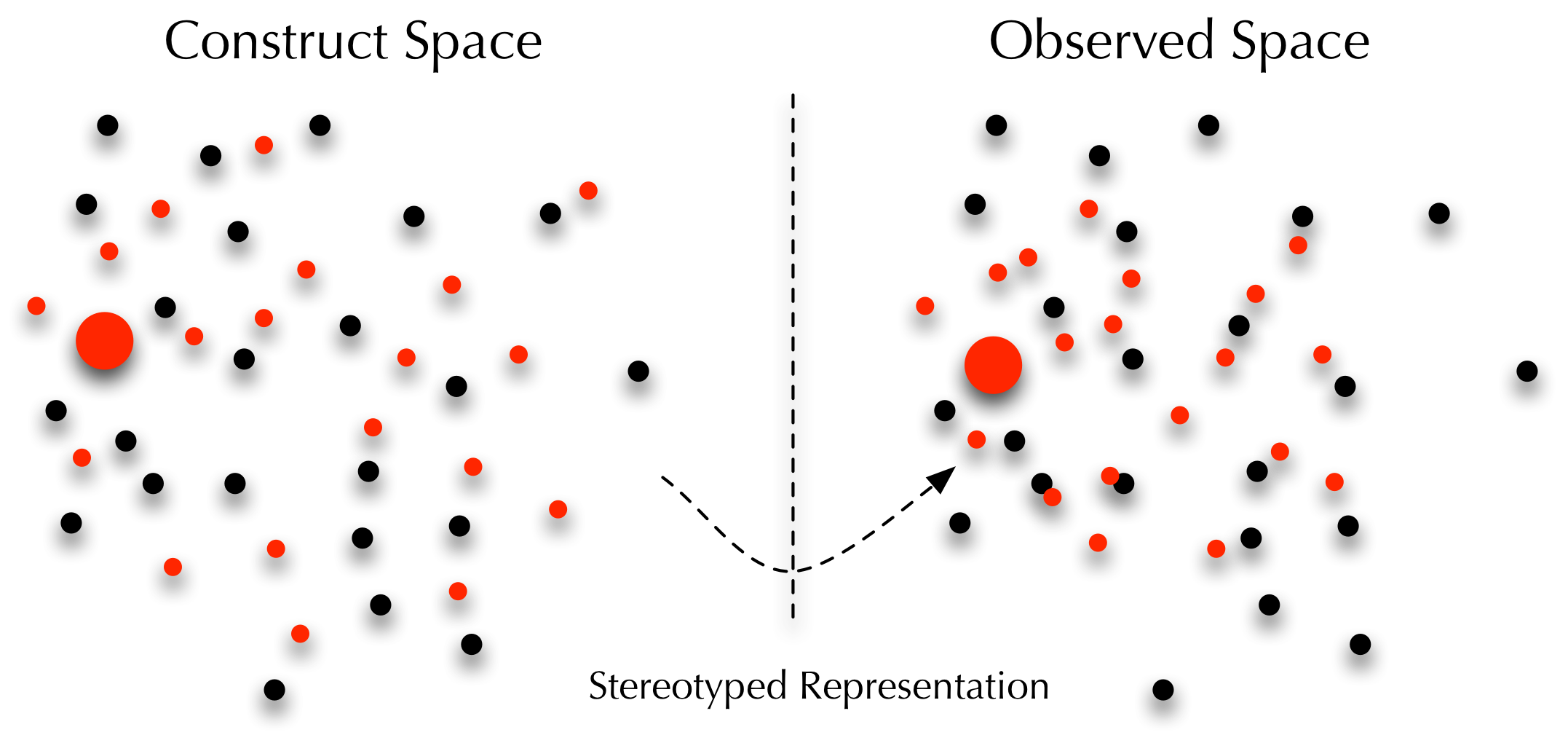}
    \caption{An example of stereotyping: points are drawn to the exemplar}%
    \label{fig:stereotype}%
\end{figure}

This mechanism is illustrated in Figure~\ref{fig:stereotype}. The left subfigure represents the original representation in the construct space with two groups colored in red and black. Once an exemplar (the larger red point) is chosen, the observation process associates other point more closely with it -- this is effectively achieved by having the red points mapped closer to the exemplar, as we see in the right subfigure.

Formally, we model this mechanism as follows. Let the minority (protected) group point set be denoted by $P$ and the majority (unprotected) group point set be $U$. Fix an exemplar $c \in P$. Then each point $p \in P$ is shifted as follows:
\begin{equation}
  p_\alpha = (1-\alpha)p + \alpha c\label{eq:7}
\end{equation}
The term $\alpha$ is the \emph{measure of stereotyping}. Note that if $\alpha = 0$, no stereotyping happens, and if $\alpha = 1$, all points are collapsed to the exemplar. Points in $U$ are not shifted at all. 

\paragraph{Stereotyping Using Features.} Stereotyping might happen only along some dimensions of the data, and not others. For example, we might stereotype all Asians as having higher aptitude for math  based on an exemplar, but we might not borrow other attributes of the exemplar (say proficiency in sports, or crafting skills) and extend them also to all Asian people. Formally, this amounts to performing stereotyping in a subspace of relevant features. Assume that only $k$ of $d$ features are influenced by stereotyping. Then we can write the mechanism for shifting a point $p \in P$ as before as
\begin{equation}
 p_\alpha = \begin{pmatrix} (1-\alpha)\mathbf{I}_k & \mathbf{0} \\ \mathbf{0}
    & \mathbf{I}_{d-k}\end{pmatrix} p + \begin{pmatrix} \alpha \mathbf{I}_k &
    \mathbf{0} \\ \mathbf{0} & \mathbf{0} \end{pmatrix} c   \label{eq:10}
\end{equation}
where we have reordered features so that the stereotyped ones are the first
$k$.

%\subsection{Stereotyping Using Features.}
%\label{sec:ster-using-feat}
%
%Stereotyping by exemplar changes all the features of a point. However, it might be that stereotyping happens only along some dimensions of the data, and not others. For example, we might stereotype all Asians as having higher aptitude for math  based on an exemplar, but we might not borrow other attributes of the exemplar (say proficiency in sports, or crafting skills) and extend them also to all Asian people. Formally, this amounts to performing stereotyping in a subspace of relevant features. Assume that only $k$ of $d$ features are influenced by stereotyping. Then we can write the mechanism for shifting a point $p \in P$ as before as
%\begin{equation}
% p_\alpha = \begin{pmatrix} (1-\alpha)\mathbf{I}_k & \mathbf{0} \\ \mathbf{0}
%    & \mathbf{I}_{d-k}\end{pmatrix} p + \begin{pmatrix} \alpha \mathbf{I}_k &
%    \mathbf{0} \\ \mathbf{0} & \mathbf{0} \end{pmatrix} c   \label{eq:10}
%\end{equation}
%where we have reordered features so that the stereotyped ones are the first
%$k$. Once again, $\alpha$ is the measure of stereotyping and ranges between $0$ and $1$. 

\subsection{Stereotyping via Representativeness.}
\label{sec:ster-feat}

We now turn to a probabilistic mechanism for stereotyping first proposed by Bordalo et al\citep{bordalo2016stereotypes}.

For ease of exposition, imagine a data representation consisting of single feature that takes $T$ distinct values that we call ``types''. For example, the feature might be \texttt{age} and the types might be specific age ranges. Let $G$ be a \emph{group} of individuals. The \emph{representativeness} of a type $t\in T$ for group $G$ is defined as the ratio of conditional probability of individuals in $G$ having value $t$ to the corresponding probability for the complement $\overline{G}$:
\[R(t,G,\overline{G}) = \frac{\Pr[t|G]}{\Pr[t|\overline{G}]}\]
Intuitively, the representativeness measures how distinctive $t$ is at distinguishing groups. The larger it is, the more likely it is that the presence of $t$ predicts group membership.

Stereotyping now occurs by \emph{amplifying} perceived probabilities based on representativeness. 
Formally, the distorted perceived probabilities are
\begin{equation}
\Pr^{st}(t|G) = \Pr(t|G)\times \frac{h_t(R(t,G,\overline{G}))}{\sum_{s\in T}\Pr[s|G]h_s(R(s,G,\overline{G}))}\label{eq:11}
\end{equation}

Where $h_t:R_+^T \rightarrow R_+$ is weakly increasing in $t$ and weakly decreasing in the other $T-1$ types.

For concreteness (and as suggested by Bordalo et al) let us assume that $h(\cdot)$ takes the form $h_t(x) = x^\rho$. Here $\rho > 1$ is again the measure of stereotyping. The larger it is, the more representativeness influences the perceived probabilities.

\subsection{A unified view of stereotyping.}
\label{sec:unif-view-ster}

While the geometric and probabilistic mechanisms described above look quite different on the surface, they are actually examples of a more general linear framework for thinking about stereotyping. For a point $p \in P$, let $v(p) : P \to F$ be an invertible transformation of $p$ into a feature space $F$. We will define a \emph{generalized stereotyping transform} parametrized by matrices $A, B$ as the transformation
$ p' = v^{-1}(A v(p) + B) $
or more conveniently
$ v(p') = A v(p) + B $
Note that setting $v$ to the identity mapping already recovers the two geometric transformations via equations \eqref{eq:7},\eqref{eq:10}. 

Consider the probabilistic transform defined by \eqref{eq:11}. As suggested, let us set $h(x) = x^\rho$. Further, let us define $\lambda(t)  = \Pr(t \mid G)/\Pr(t \mid \overline{G})$. We can now rewrite \eqref{eq:11}.

\begin{equation}
   p'_{t|G} = \frac{p_{t|G}\lambda(t)^\rho}{\sum_s p_{s|G}\lambda(s)^\rho}\label{eq:1}
\end{equation}

By assumption, probabilities in the majority group $\overline{G}$ are not
modified, which implies that $p'_{t | \overline{G}} = p_{t |
  \overline{G}}$. Dividing both sides of \eqref{eq:1} by the left and right
sides of this equality, we get
\begin{align}
  \label{eq:2}
  \lambda'(t) &= \frac{\lambda(t)^{1+\rho}}{\sum_s p_{s|G}\lambda(s)^\rho} \\
  \intertext{which after taking logs yields}
  \ln \lambda'(t) &= (1+\rho)\ln \lambda(t) - \ln \sum_s p_{s|G}\lambda(s)^\rho
\end{align}

Setting $v(p) = \ln p$ and noting that $\ln \lambda(t) = \ln p_{t|G} - \ln p_{t|\overline{G}}$ where $p_{t|\overline{G}}$ is fixed, we can recover the same linear relationship as before.

While we can unify the different mechanisms of stereotyping mathematically, we
observe that the actual processes by which these mechanisms might modify data
are different. Representativeness is a form of cognitive bias that would
manifest itself in data collection: for example in predictive policing, officers
that might stereotype potential criminals based on race are more likely to
observe (and therefore collect data) on crimes perpetuated by
minorities. Stereotyping via exemplars (and features) might manifest itself at
the time of feature selection: by omitting key differentiating features in the
representation, points might end up appearing closer than they actually are. 
%%% Local Variables:
%%% mode: latex
%%% TeX-master: "stereo"
%%% End:

\section{Harms of Stereotyping}
\label{sec:classification}

While we believe that stereotypes are harms independent of their implications for the machine learning process, the use of stereotyped data representations as training data additionally leads to disparities in outcomes by group.  In this section, we demonstrate this analytically for Naive Bayes classification, linear regression, and clustering;  thus demonstrating these harms in classification, regression, and unsupervised settings.  For Naive Bayes classification, we adopt the probabilistic interpretation of stereotyping, while the other two are studied under the geometric interpretation.

\subsection{Naive Bayes Classifier.}
\label{NaiveB}
For an arbitrary input $X$, Bayes' rule states that the probability of holding a class label $C_k, 1 \leq k \leq m$ given $X$ is:
$\Pr[C_k|X] = \frac{\Pr[X|C_k]\Pr[C_k]}{\Pr[X]}$
The goal in classification is to predict the most likely label for a given
input. Naive Bayes classification makes the assumption that conditioned on the
class label, these features are independent of each other, yielding the
following predictive rule.
$\hat{y} = \argmax_k \Pr[C_k] \prod_{i=1}^n \Pr[x_i|C_k]$
where $x_i, 1\leq i \leq n$, is a single feature of $X$.
In the presence of stereotyping such an assumption might have an undesired
impact on the classification results. Consider a dataset with sensitive
attribute $A$, e.g., being \emph{Asian}, and an attributes $D_p$ relevant to
known stereotypes for $A$, e.g. being \emph{good at math}. Let's assume $D_p$ is \emph{positively} correlated with being Asian. In this setting, if the data gathering process is susceptible to this type of stereotyping, the Naive Bayes formulation would turn into:
\begin{multline}
\hat{y} = \argmax_k \Pr[C_k] \prod_{i=1}^{n-3}\Pr[x_i|C_k] \times \\
\Pr[D_p|A,C_k]\Pr[A|C_k]
\end{multline}
where $\Pr[D_p|C_k]$ is replaced by $\Pr[D_p|A,C_k]$. According to the probabilistic interpretation of
stereotyping \citep{bordalo2016stereotypes}, the perceived probability
$\Pr[D_p|A]$ for the target group could be different from its actual value:
$\Pr^{st}[D_p|A,C_k] \geq \Pr[D_p|A,C_k]$
which can influence the classification results. Indeed, the degree of difference
is related directly to the representativeness as a function of the ratio
$\Pr^{st}[D_p|A, C_k]/\Pr[D_p|A, C_k]$.

This transition -- from $\Pr[D_p| C_k]$ to $\Pr[D_p | A, C_k]$ is key to
understanding stereotyping in this context. In other words, the point of a
predictive tool is that it uses ``other'' features other than the class label to
determine the probabilities for desirable attributes like $D_p$. Alternatively,
we can think of $\Pr[D_p|A, C_k]$ as reflecting a process by which the variable
$A$ \emph{affects} the data used to compute the conditional probability, either
by removing data that does not have $A$ or by overweighting data that does.

\subsection{Linear Regression.}
\label{sec:linear}
In linear regression, we are given points $X = x_1, \ldots, x_n \in \reals^d$ and
corresponding labels $y_1, \ldots, y_n \in \reals$. The goal is to find
a vector of parameters $\beta$ such that $y = X\beta + \epsilon$.
It is well known that the least-squares solution to this regression is given by 
$\beta = (X^\top X)^{-1}X^\top y$.
and our goal is to understand how perturbing the input $X$ (via stereotyping)
will change the coefficients\footnote{There is an extensive literature on the
  stability of linear regression to \emph{random} perturbations of the input\cite{chatterjee2009sensitivity};
  the behavior of the coefficients is well understood for example when the
  inputs are perturbed using Gaussian noise. In our setting, the perturbations
  are not random and are structured in a specific way, and thus the prior forms
  of analysis, and even more recent work that looks at other forms of structured
  perturbations\cite{diaz2007exact} appear
  to not apply directly.}.

We will consider a very simple form of perturbation, where only a single
coordinate $s$ is perturbed while the rest (including the dependent variable $y$) stay fixed.
 Assume that the data matrix $X$ (in which each
point is a row) is organized as $\begin{pmatrix}  Q \\ P\end{pmatrix}$ where $Q$
is the set of all majority group points and $P$ consists of all minority group
points. Let $c$ be the exemplar that points are pulled towards (in dimension
$s$). We can then write the perturbation as
\[X' = X + \alpha(\mathbf{\Xi}{c^*}^\top - I_rXI_s)\]
where $\mathbf{\Xi}$ is a $n \times 1$ vector with values of 1 for minority data
points and zeros for majority ones and $c^*$ is a vector of all zeros except in
the $s^{th}$ position, where its value is $c_s$.
% \[{c^*}_{d \times 1}:{c^*}_i = \begin{cases}
%     c_i,& \text{if } i=s\\
%     0,& \text{otherwise}
% \end{cases}\]
% represents the value towards which all the minority data points are pulled, in the perturbed dimension;
$I_r$ is an $n \times n$ diagonal matrix with values of 1 for rows in $X$ representing minority data points and zeros everywhere else; $I_s$ is a $d \times d$ matrix where its only non-zero element, which is 1, is at row and column $s$.

For a general perturbation of the form $X' = X + \Delta$
the coefficients in $\beta$ are updated to:
\begin{equation}
\label{coefeq}
\beta' = [(X+\Delta)^\top (X+\Delta)]^{-1}(X+\Delta)^\top y
\end{equation}
The key term in the right hand side is the inverse, which can be written in
general form as $[X^\top X + (\Delta^\top \Delta + X^\top \Delta + \Delta^\top
X)]^{-1}$. This is asking for the inverse of a perturbation of a given matrix,
for which we can use the famous Woodbury formula\cite{woodbury}:
\begin{equation}
(A + UCV)^{-1} = A^{-1} - A^{-1}U(C^{-1}+VA^{-1}U)VA^{-1}) \notag
\end{equation}

Let the number of
rows in $P$ (the number of minority points) be $m$ and $\mu$ be the centroid of
the points in $P$. We can then write
\begin{equation}
\begin{split}
&X'^\top X' = \\
&X^\top X + \alpha(m\mu {c^*}^\top - P^\top PI_s) + \alpha(m c^* \mu^\top - I_sP^\top P)\\
&- m\alpha^2(c^* \mu^\top I_s + I_s\mu {c^*}^\top + c^* {c^*}^\top) + \alpha^2
I_sP^\top PI_s \notag
\end{split}
\end{equation}
Therefore, the update to $X^\top X$ consists of two of rank-1 updates:
$X'^\top X' = X^\top X + wu^\top + uw^\top$
where:
\noindent\resizebox{\columnwidth}{!}{
$w_i = \begin{cases}
    \alpha (mc\mu_s - \Vert P^\top _s\Vert^2) + \frac{m\alpha^2(c^2-2c\mu_s) + \alpha^2 \Vert P^\top _s\Vert^2}{2},& \text{if } i=s\\
    \alpha (mc\mu_i - P^\top _i \cdot P^\top _s),              & \text{otherwise}
  \end{cases}
$
}
and $u$ is a basis vector with a $1$ in dimension $s$. Setting 
%Having the vectors for rank-1 updates defined, decomposition of the update matrix, $UCV$, can be written as:
$U_{d \times 2} = \begin{pmatrix}w & u\end{pmatrix},V_{2 \times d}
= \begin{pmatrix}u \\w\end{pmatrix},C = I_2$ and embedding these matrices in the Woodbury formula yields
\begin{equation}
\label{eq:newbeta}
\beta' = (\underbrace{X^\top X)^{-1} X'^\top Y}_{p_1} - \underbrace{(X^\top X)^{-1} M (X^\top X)^{-1}X'^\top Y}_{p_2}
\end{equation}
The amount of perturbation is controlled by $\alpha$. If we split Equation
\ref{eq:newbeta} into two parts as illustrated, the values in part $p_1$ change
linearly in $\alpha$ because of the linear dependence on $X'$. However, each
element in $M$ is a quadratic function of $\alpha$. Therefore, the values in
part $p_2$, change quadratically when $\alpha$ is increased. % For example, the
% values for $\beta$ coefficients based on the value of $\alpha$, are illustrated
% in Figure \ref{fig:lrexperiment} part (c), for the case where the lowest value
% in the stereotyping dimension is chosen as the landmark. \sverror{do we have
%   this still?}

\subsection{Clustering.}
\label{sec:cluster}

Unlike in the previous two cases, we will not present a formal analysis of how
clustering is affected by stereotyping, because clustering (and especially
$k$-means) is a global objective where it is often difficult to predict how
perturbations will affect the outcome. Rather, we will demonstrate empirically
the effect of stereotyping on clustering in Section \ref{sec:experiments-1}.

However, we argue qualitatively here for why clustering will be affected by
stereotyping. The effect of moving points closer to an arbitrarily chosen
exemplar, especially if this exemplar is not the mean of the set of minority
points, has the effect of shifting and concentrating clusters to make them look
more homogeneous with respect to group identity. But if the exemplar is an
outlier, then the cost of the clustering will increase: for $k$-means this
increase in cost is related both to the distance between the exemplar and the
cluster mean and the number of points that are moved. 

%%% Local Variables:
%%% mode: latex
%%% TeX-master: "stereo"
%%% End:

\section{Mitigating the Effects of Stereotyping}
\label{sec:mitigation}
Since it is impossible to access construct space, in order to mitigate the
unwanted influences of stereotyping on machine learning pipeline, we need to
make assumptions about data in the ideal world. One useful assumption to make is
based on \emph{We're All Equal} worldview \citep{friedler2016possibility},
where the idea is that in the ideal world, different groups look essentially the same. 
Though such assumption might not hold true in every possible scenario (e.g. women are on average
shorter than men), it implicitly appears in much of literature on statistical discrimination.
This motivates us to adopt \emph{WAE} as an appropriate axiom in our work. Let's consider 
two social groups in a hypothetical dataset: minority and majority. The minority 
group is the one being stereotyped in the mapping from construct to observed space. 
Based on the \emph{WAE} worldview, we assume the two groups are generated by same
distribution which we can estimate, by looking at the majority group in the
observed space. Therefore, the goal is to recover the true representation of the
minority group in the observed space, based on majority group.

\subsection{Mitigation of exemplar-based stereotyping.}
\label{sec:mitig-under-geom}

Recall that in landmark-based stereotyping, a landmark $c$ is fixed first, as
well as a measure of stereotyping $\alpha$. Then each point in the minority group is pulled towards
the landmark resulting in a new point $p_\alpha$.
Let the resulting mean of the modified points be $\mu_\alpha$.  Using
equation~\eqref{eq:7} and noting that the
stereotyping process is linear, we can write
\begin{equation}
\mu_\alpha = (1-\alpha)\mu_m + \alpha c\label{eq:6}
\end{equation}

where $\mu_m$ is the mean of all the points in the minority group. Our goal is
now to determine the values of $\alpha$ and $c$ so that we can reconstruct the
original points. We will invoke the WAE assumption by assuming that the mean $\mu_m$ is close to
the mean $\mu_M$ of the majority group: specifically, that $\|\mu_m - \mu_M\|
\le \epsilon$. 

Our goal is to determine candidates for $c, \alpha$. \eqref{eq:6} tells us that
any feasible exemplar $c$ must lie on the line between $\mu_\alpha$ and
$\mu_m$. Since by assumption $\mu_m$ lies in a ball of radius $\epsilon$ around
$\mu_M$ (denoted by $B_\epsilon(\mu_M)$), the feasible region for $c$ is a cone with apex
at $\mu_\alpha$ such that the surface of its complementary cone is tangent to
the ball around $\mu_M$. Let $\|\mu_\alpha - \mu_M\| = d$, and fix a point
$c$. Then $c$ is a feasible exemplar if the angle $\theta$ made by $c - \mu_\alpha$ with
the vector $\mu_\alpha - \mu_m$ is such that $\sin\theta \le \epsilon/d$. We
denote the set of such vectors by $C_{\mu_\alpha, \mu_M}(\epsilon)$ and
will drop the subscripts for notational ease.

We have assumed that the exemplar is a point in $P$, so the set of feasible
exemplars can be written as $P \cap C(\epsilon)$. For each such $c$
the set of possible locations for $\mu_m$ is a line segment resulting from the intersection of
the line supporting the segment $\overline{\mu_\alpha c}$ and
$B_\epsilon(\mu_M)$. For each candidate choice of $\mu_m$ on this segment we can
compute $\alpha$ using \eqref{eq:6}. We will seek the smallest possible $\alpha$
-- this represents the most conservative measure of stereotyping that is
consistent with the observed data. It is easy to see that this minimum is
achieved by the endpoint of the line segment closest to $\mu_\alpha$ and can be
computed easily.

\begin{figure}[htbp]
  \centering
  \includegraphics[width=0.8\columnwidth]{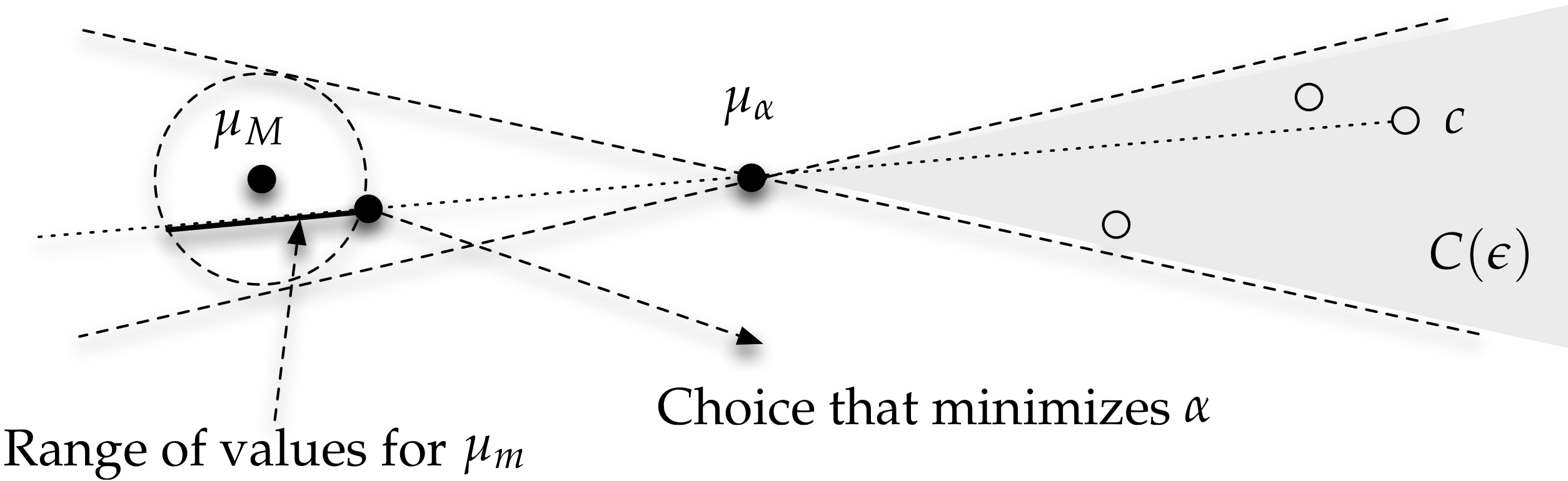}
  \caption{How we compute candidate $c, \alpha$ values}
  \label{fig:landmark}
\end{figure}

We now can associate a value of $\alpha$ with each $c \in P \cap
C(\epsilon)$. We take the pair $(c,\alpha)$ such that $\alpha$ is
minimized. This represents the most conservative reconstruction that is
consistent with the WAE assumption and the observed data. We summarize this
process as follows:

\bignote{\small\begin{enumerate}
\item Compute the set $P \cap C(\epsilon)$ by verifying the angle test for each
  point of $P$. 
\item For each candidate $c$ in this set, compute its associated value of $\alpha$. 
\item Return the pair $(c, \alpha)$ that minimizes $\alpha$. 
\end{enumerate}
}
\subsubsection{Mitigating stereotyping with respect to features.}
\label{sec:mitig-ster-with}

The process above captures stereotyping with respect to exemplar
points. We now consider the case of mitigating stereotypes that are limited to
some features. As before, we can exploit the linearity of the perturbation \eqref{eq:10}
and the same idea as in the
previous subsection to observe that the means $\mu_\alpha, \mu_m$ satisfy the same
relationship. 
\[ \mu_\alpha = \begin{pmatrix} (1-\alpha)\mathbf{I}_k & \mathbf{0} \\ \mathbf{0}
    & \mathbf{I}_{d-k}\end{pmatrix} \mu_m + \begin{pmatrix} \alpha \mathbf{I}_k &
    \mathbf{0} \\ \mathbf{0} & \mathbf{0} \end{pmatrix} c \]

This reduces to a $k$-dimensional version of the full stereotyping problem
described above.%  where the search process operates in $k$ dimensions but returns
% points from $P$.

\subsection{Mitigating Representativeness.}
\label{sec:mitig-repr}

% Recall equation \eqref{eq:2} for stereotyping derived in
% Section~\ref{sec:unif-view-ster}.

%   % \lambda'(t) &= \frac{\lambda(t)^{1+\rho}}{\sum_s p_{s|G}\lambda(s)^\rho} \notag\\
%   % \intertext{which after taking logs yields}
% $  \ln \lambda'(t) = (1+\rho)\ln \lambda(t) - \ln \sum_s p_{s|G}\lambda(s)^\rho $

We introduce the WAE modeling assumption: we assume that prior to
stereotyping, the probabilities of types are very similar between majority and
minority groups. Formally, we will express this condition as
$  1-\epsilon \le \lambda(t)  \le 1+\epsilon$
where $\epsilon$ is a parameter that controls the degree to which the
distributions are similar. Note that this assumption implies that the
Kullback-Leibler divergence between the two conditional  distributions is at most
$\epsilon$. This follows from the fact that the KL-divergence can be written as
$ d_{KL}(\mathcal{D_G}, \mathcal{D_{\overline{G}}}) = \sum_t \Pr(t | G)\ln \lambda(t)$.
% \begin{align}
%  d_{KL}(\mathcal{D_G}, \mathcal{D_{\overline{G}}}) &= \sum_t \Pr(t | G)\ln \lambda(t)  \\
%                                                  &\le \sum_t \Pr(t | G)\ln (1+\epsilon) \\
%                                                  &\le \epsilon \sum_t \Pr(t | G)\\
%   &= \epsilon
% \end{align}
Note that $ \sum_s p_{s|G}\lambda(s)^\rho$ is a convex combination of the
quantities $\{ \lambda^\rho(s) \}$. It will be convenient to express this sum as
$e^{\gamma\rho}$ where $e^\gamma \in [1-\epsilon, 1+\epsilon]$. Setting
$\delta_t = \ln \lambda(t)$, we note that by the standard approximation of
$\ln(1+x)$, $|\delta_t| \le \epsilon$. 

We can now substitute these bounds on $\lambda(t)$ into \eqref{eq:2}, yielding
  $\ln \lambda'(t) = (1+\rho)\delta_t - \gamma \rho = \delta_t + \rho(\delta_t - \gamma_t) $
where $|\delta_t|, |\gamma_t| \le \epsilon$.
Solving for $\rho$ yields
$  \rho = \frac{\ln \lambda'(t) - \delta_t}{\delta_t - \gamma}$.
There is one such equation for $\rho$ for each value of $t$. The terms
$\delta_t$ and $\gamma$ are unknown (since they arise from the (unknown)
distribution of types prior to stereotyping). However, we can use the bounds on
these terms to provide bounds on the value of $\rho$.

The minimum value of $\rho$ implied by any of the equations can be determined by
setting $\delta_t$ to its largest value and $\gamma$ to its smallest. This
yields $\rho = \frac{\ln \lambda'(t) - \epsilon}{2\epsilon}$. Note that $\rho$
can grow without bound, which implies that each equation yields a half-infinite
range of possible values of $\rho$. Taking the intersection of all these ranges
we conclude that $\rho$ must lie in the range $[\frac{\ln \max \lambda'(t) -
  \epsilon}{2\epsilon}, \infty]$.

Each value of $\rho$ in this range is a candidate measure of stereotyping and
can be used to reconstruct the original probabilities $p(t | G)$. Specifically,
it is easy to see that $\lambda(t)$ is proportional to $\lambda'(t)^{1/\rho}$,
and since we can compute $p(t|\overline{G})$ directly we can obtain
$p(t|G)$. Interestingly, if we now compute the KL-divergence between
$\mathcal{D}_G$ and $\mathcal{D}_{\overline{G}}$, it is a monotonically
decreasing function of $\rho$. In other words, the smallest feasible value of
$\rho$ given above is consistent with the WAE assumption and is also the most
conservative choice. % (in that the effect of stereotyping is taken to be as small
% as possible under the WAE assumption).

Summarizing, our procedure is:
\vspace*{-0.1in}
\bignote{\small\begin{enumerate}
\item Compute $\lambda'(t)$ for all types $t$. 
\item Set $\rho = \frac{\ln \max \lambda'(t) - \epsilon}{2\epsilon}$
\item Set $p_t = \lambda'(t)^{1/\rho}$.
\item Set $\tilde{p}(t |G ) = \frac{p_t}{\sum_t p_t}$. Return $\{p(t|G)\}$. 
\end{enumerate}
}

%%% Local Variables:
%%% mode: latex
%%% TeX-master: "stereo"
%%% End:

\section{Experiments}
\label{sec:experiments-1}

In this section, we provide an empirical assessment for harms of stereotyping to Naive Bayes classification, linear regression and clustering, using synthetic datasets. Also, for each one of these problems, we demonstrate the effectiveness of our mitigation methods. 

\subsection{Naive Bayes.}
\label{sec:nb-experiment}

In order to study the impact of stereotyping via representativeness we consider a synthetic dataset that allows us to manipulate the extent of stereotyping in the data and note its effect on a Naive Bayes classifier.  The synthetic data contains three binary attributes and a class label: \begin{enumerate*}
\item the \emph{sensitive} attribute which is assigned randomly and specifies if an individual is Asian or not;
\item a randomly assigned attribute that has no correlation with the class label or sensitive attribute;
\item an attribute indicating if the individual is \emph{good at math}, which contributes positively to receiving the desired classification; and
\item a class label indicating if the individual is \emph{selected} for a job interview. 
\end{enumerate*} 
2000 instances are generated according to this procedure with a 50:50 training to test split ratio. 

Stereotyping via representativeness can be thought of as amplifying disparities that already exist between target and base subgroups.  The extent of these disparities is measured by $\lambda(t)$, 
%We set $\lambda(t)$ to be the conditional probability of being ``good at math" given an individual is Asian, to that if they were not, 
and we assume that $\lambda(t) > 1$ for the type ``good at math" with respect to the Asian subgroup.  In order to simulate a stereotyped dataset, first we fix $\lambda(t)$ by assigning positive values to ``good at math" for Asians, with a higher probability compared to non-Asians. Then, we gradually increase the positive correlation between being Asian and the attribute ``good at math."  Specifically, the integer value of $\rho$ is increased from 1 to 10 where
\begin{equation}
p'_{t|G} = \frac{p_{t|G}\lambda(t)^\rho}{\sum_s p_{s|G}\lambda(s)^\rho}
\end{equation}
given that $\lambda(t)  = \Pr(t \mid G)/\Pr(t \mid \overline{G})$ and $G$ and $\overline{G}$ denote Asian and non-Asian groups respectively.

\begin{figure}[htbp]%
    \centering
%    \subfloat[Number of selected Asians increases with $\rho$.]{\includegraphics[width=0.45\columnwidth]{nbselected.pdf} \label{fig:nbstereo}}%
%~~~~
    \subfloat[Number of selected Asians for different values of $\lambda$ and $\rho$]{{\includegraphics[width=0.45\columnwidth]{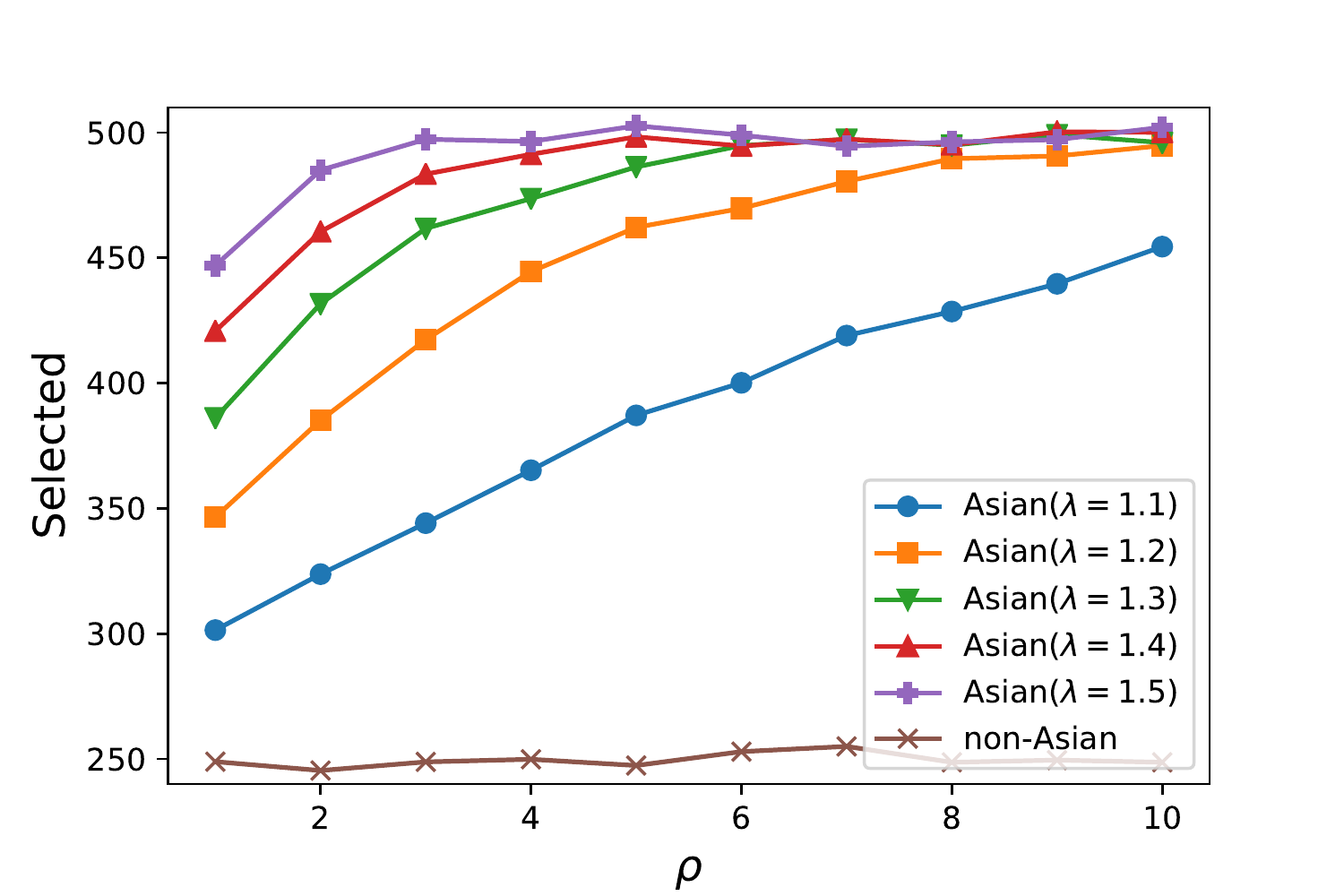} }\label{fig:nbstereo}}
    ~~~~
%    \subfloat[Number of selected Asians remains roughly the same by mitigating representativeness. ]{\includegraphics[width=0.45\columnwidth]{mit-nbselected.pdf} \label{fig:nbfix}}%
    \subfloat[Effects of mitigation on the number of selected Asians]{{\includegraphics[width=0.45\columnwidth]{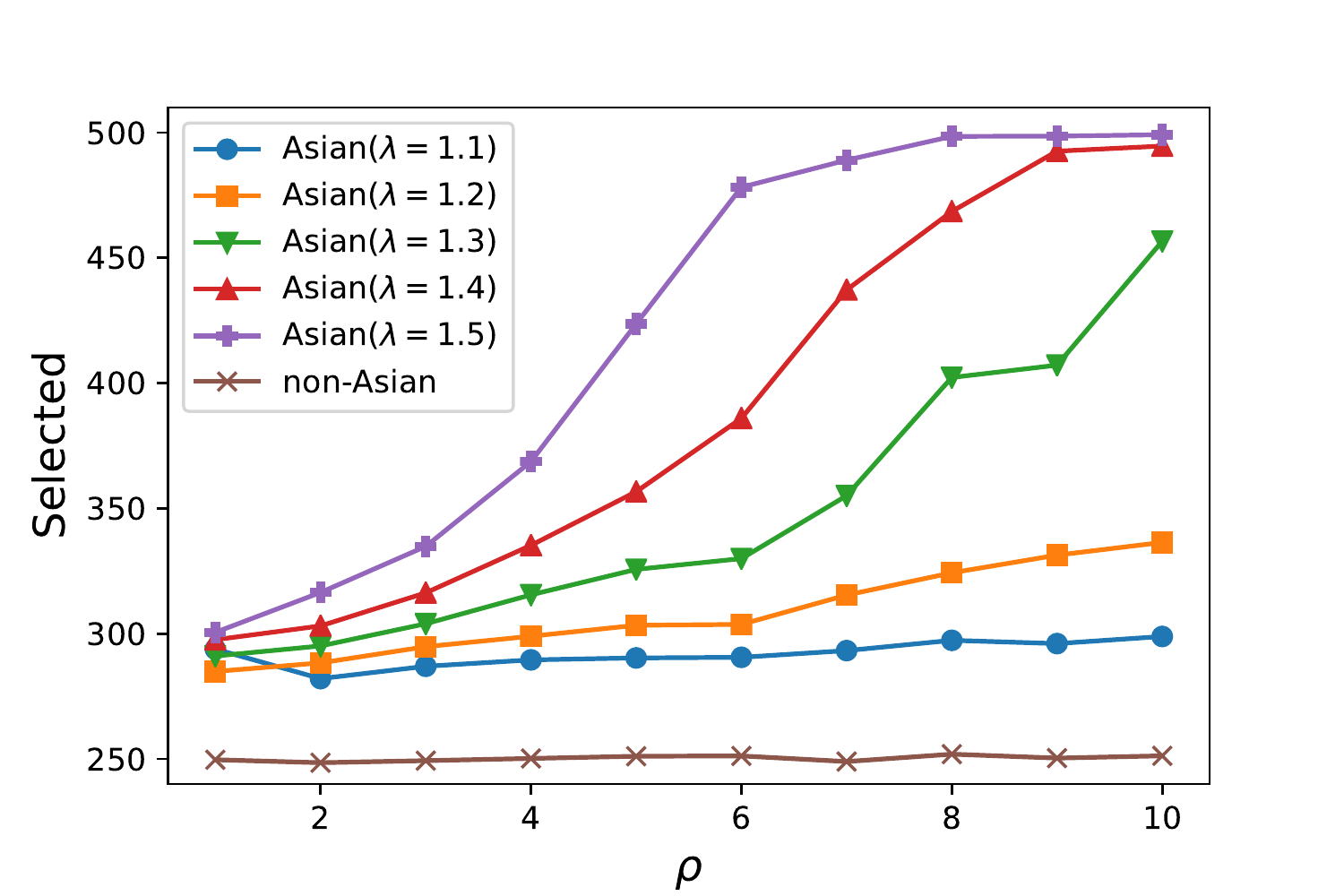} }\label{fig:nbfix}}
    \caption{Results of NB classification as the Asian subgroup's representative type ``good at math" is boosted.}%
    \label{fig:nbexperiment}%
\end{figure}

The results are shown in Figure \ref{fig:nbexperiment}. As illustrated in Figure \ref{fig:nbstereo}, by boosting the representativeness of being ``good at math," the number of selected Asians increase while no difference is observed in the other group's results. These results hold over different values of $\lambda(t)$.  Figure \ref{fig:nbfix} shows that the effects of stereotyping on the Asian subgroup's results are significantly reduced by applying the representativeness mitigation solution. We should note there is a possibility for large values of $\lambda(t)$ and $\rho$, to saturate the probability of type $t$ for the target group, e.g. $\lambda= 1.5, \rho = 10$ in Figure \ref{fig:nbfix}. In such cases, since the stereotyped probabilities for the other types go down to zero, our mitigation method would not be able to retrieve the original probabilities.

\subsection{Linear Regression.}
\label{sec:lr-experiment}
In this experiment, we study the effects of stereotyping under the geometric interpretation on linear regression.  We again consider this on a synthetic dataset so that we can manipulate the extent of the stereotyping.  The dataset contains four features: 
the first is a uniform randomly assigned binary \emph{sensitive} attribute with values privileged and unprivileged;
the second, third, and fourth features are numerical values between 0 and 1 assigned via a uniform distribution; and
the dependent variable is a linear combination of the third and fourth attributes with noise $-0.1 \leq \epsilon \leq 0.1$  added, i.e.:
\begin{equation}
y = -x_3 + 2x_4 + \epsilon \label{eq:reg}.
\end{equation}
2000 instances are generated according to this procedure, and a 50:50 training to test split ratio is used.

We assume higher values for $y$ are desired by individuals. In this experiment, we pick the individual with the lowest value for its dependent variable $y$ as the exemplar, and modify values $x_2$, $x_3$, and $x_4$ for individuals from the unprivileged group so that the distance between those individuals and the exemplar is decreased according to parameter $0<\alpha<1$ i.e. if $\alpha = 0$ the values don't change and if $\alpha = 1$ the values for $x_2$, $x_3$, and $x_4$ are the same as for the exemplar point.

\begin{figure}[htbp]%
    \centering
    \subfloat[Linear decrease in $\hat{Y}$ for unprivileged group as $\alpha$ increases]{{\includegraphics[width=0.45\columnwidth]{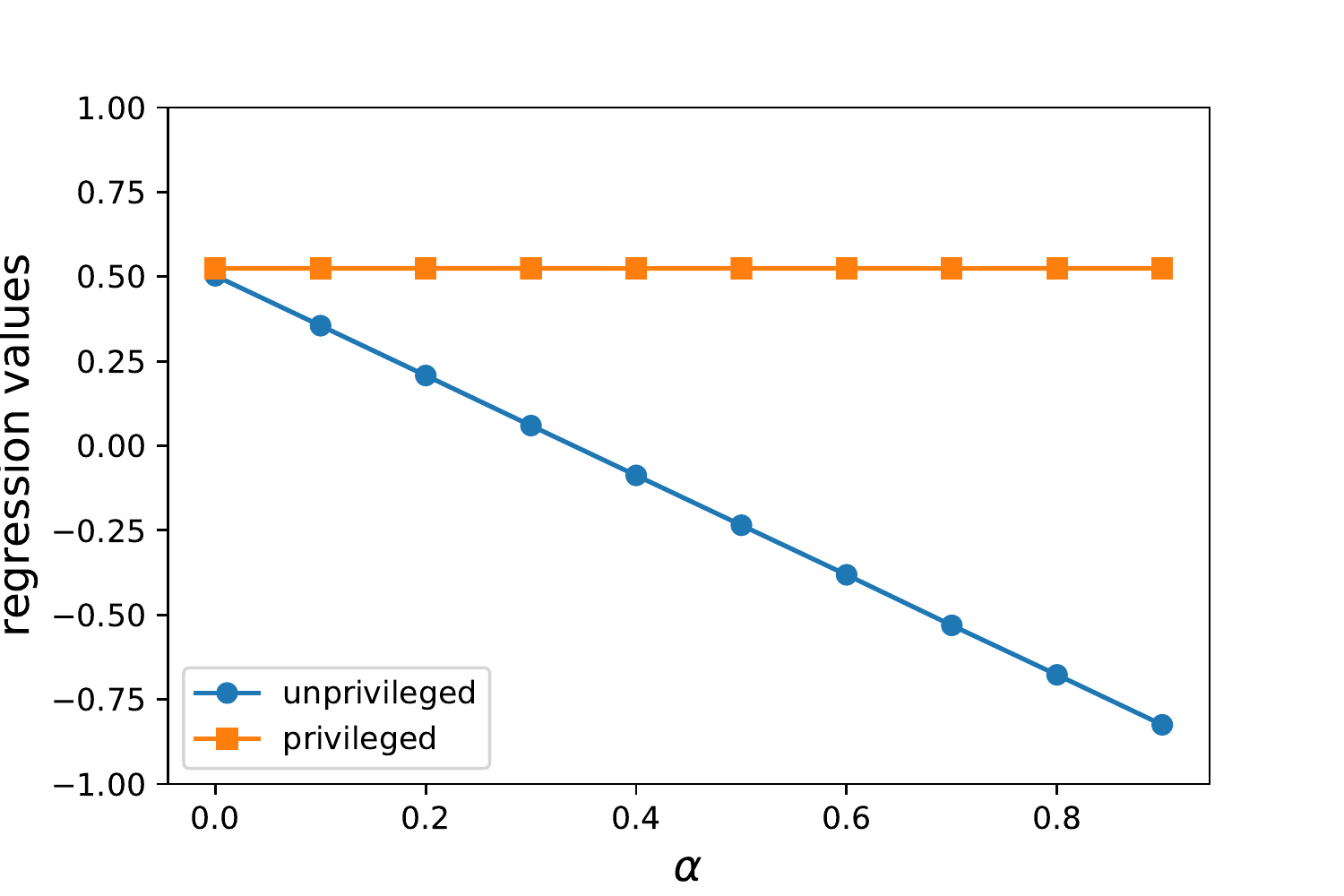} }\label{fig:lr-regs}}%
    ~~~~
    %\subfloat[The average regression values of $Y$ for both groups change quadratically with $\alpha$.]{{\includegraphics[width=0.45\columnwidth]{regs-not.pdf} }}\\
	\subfloat[The average regression values after applying mitigation]
{{\includegraphics[width=0.45\columnwidth]{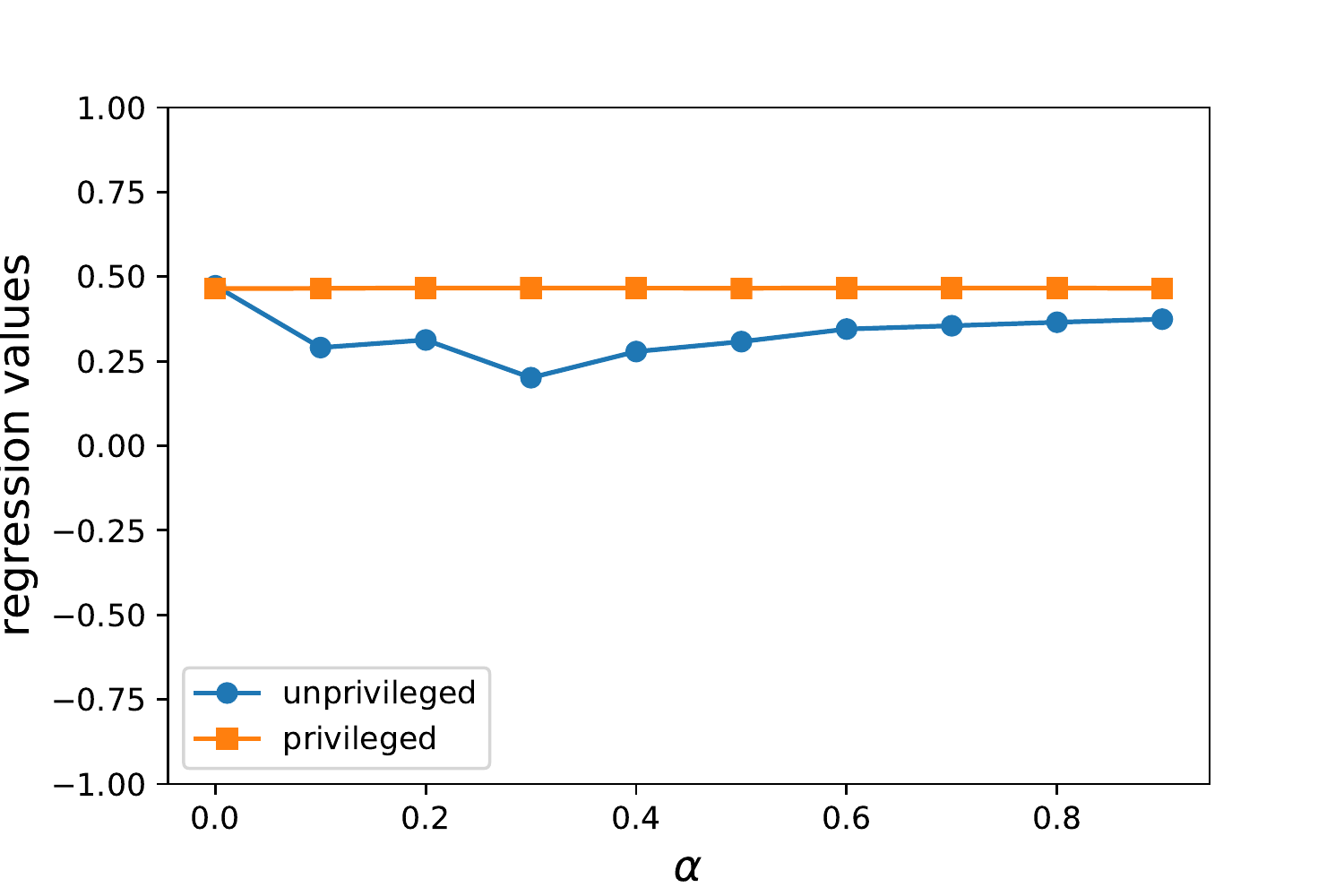} }\label{fig:mit-lr-regs}}%
    %\subfloat[The average regression values after applying mitigation by landmarks, regression values updated]{{\includegraphics[width=0.45\columnwidth]{mit-regs-not.pdf} }}    
    \caption{Changes in the regression values for the two groups as stereotyping gets more aggressive}%
    \label{fig:lrexperiment}%
\end{figure}

The results of gradually increasing the value of $\alpha$ are shown in Figure \ref{fig:lrexperiment}. In the stereotyping process, since the values for the dependent variable are updated according to regression function \ref{eq:reg}, the regression coefficients stay the roughly same. But as a result of stereotyping, there will be a disparity in the regression values as shown in Figure \ref{fig:lr-regs}. Looking at Figure \ref{fig:mit-lr-regs}, we observe that the disparities in regression values, which were caused by stereotyping, are reduced by applying the exemplar-based mitigation method.

\subsection{Clustering.}
\label{sec:clustering-experiment}
\begin{figure}[htbp]%
    \centering
    \subfloat[$\alpha = 0$]{{\includegraphics[width=0.45\columnwidth]{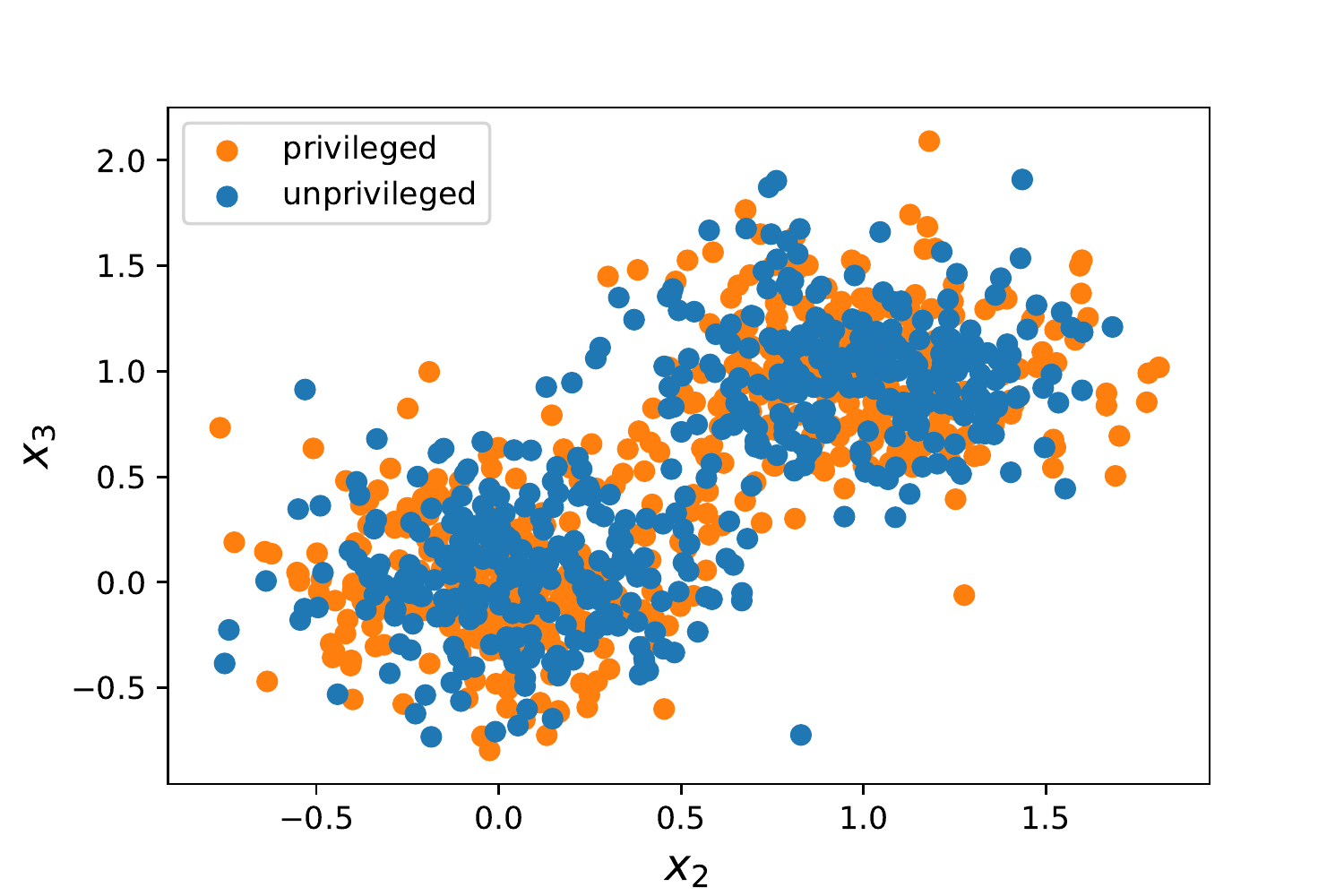} }}
    \subfloat[$\alpha = 0.3$]{{\includegraphics[width=0.45\columnwidth]{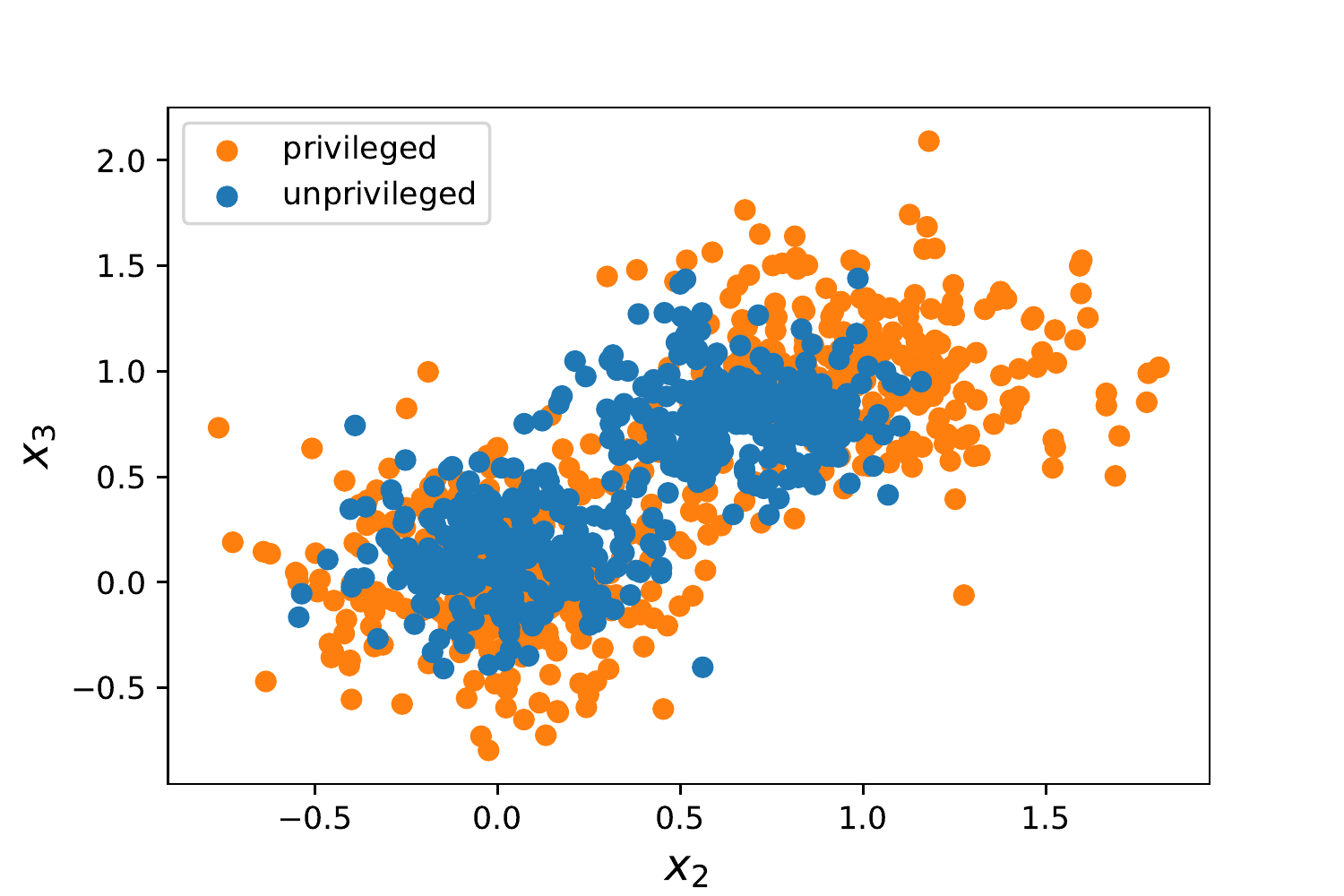} }}\\
    \subfloat[$\alpha = 0.6$]{{\includegraphics[width=0.45\columnwidth]{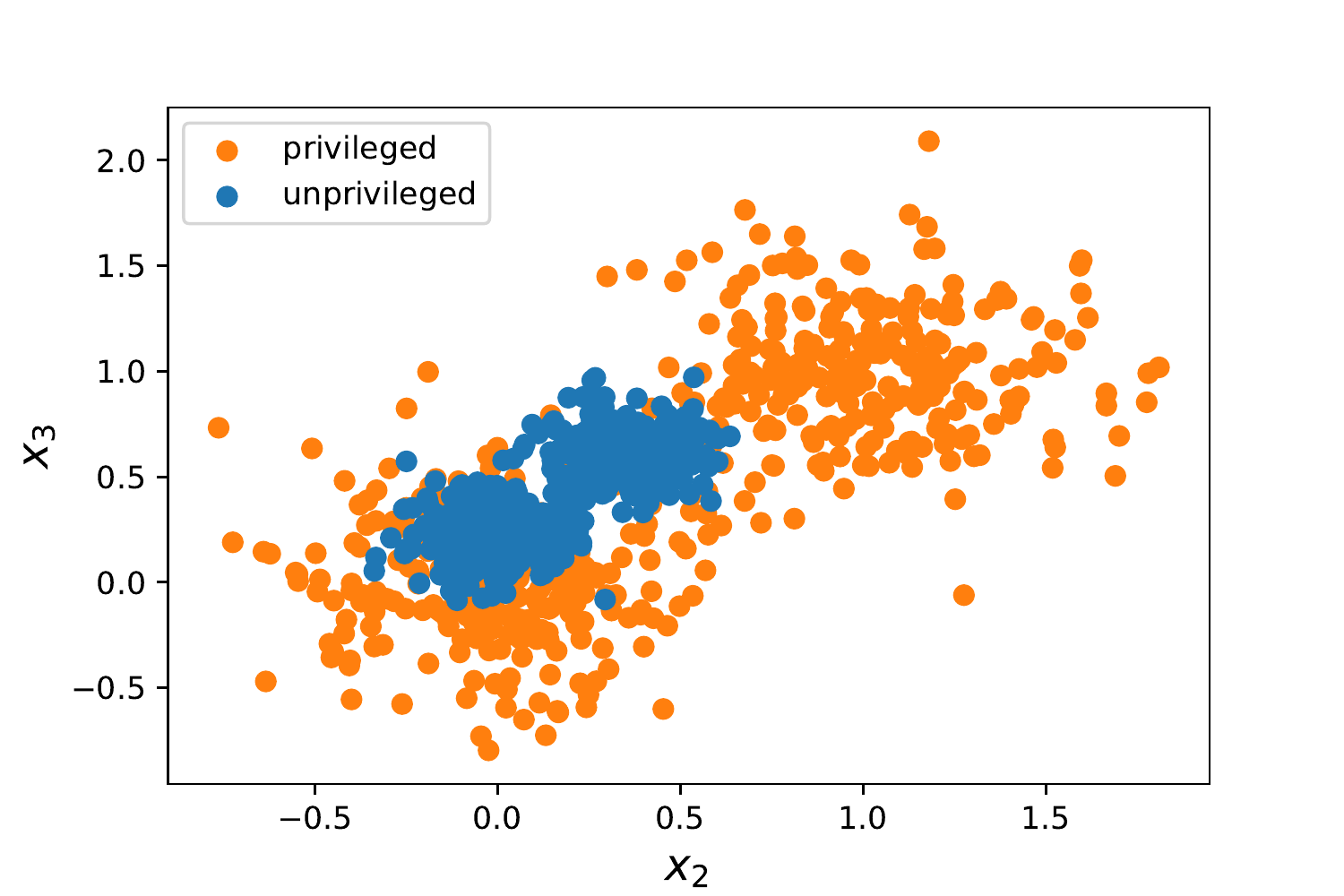} }}
    \subfloat[$\alpha = 0.9$]{{\includegraphics[width=0.45\columnwidth]{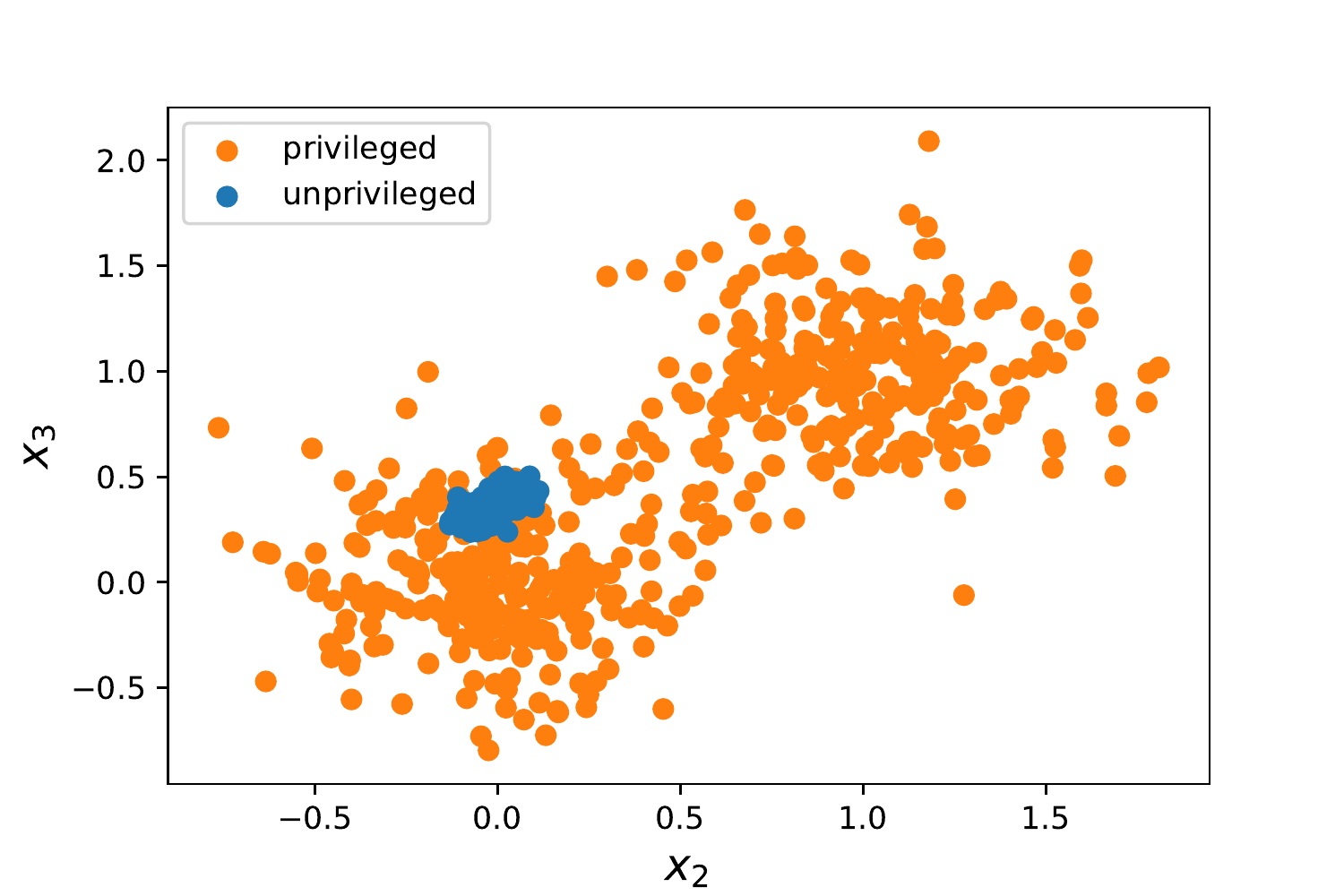} }}\\
    \caption{Stereotyping under geometrical interpretation for different values of $\alpha$}%
    \label{fig:cl-st-scatter}%
\end{figure}

We now empirically study the effects of stereotyping under geometric interpretation on $K$-means clustering. We create a synthetic dataset with three features. The first feature is a uniform randomly assigned binary \emph{sensitive} attribute with values ``privileged" and ``unprivileged". We use two 2-dimensional normal distributions $\mathcal{N}(0,0.3^2)$ and $\mathcal{N}(1,0.3^2)$
to assign values to the second and third attributes; more specifically, in each social group, the second and third attributes for  half the data are generated using one distribution and for the other half from the other. As for stereotyping, we pick an arbitrary exemplar within the unprivileged group and move the points representing the remaining individuals in this group towards it according to parameter $0<\alpha<1$, as illustrated in Figure \ref{fig:cl-st-scatter}.

\begin{figure}[htbp]%
    \centering
    \subfloat[By boosting $\alpha$, Rand index decreases.]{\includegraphics[width=0.45\columnwidth]{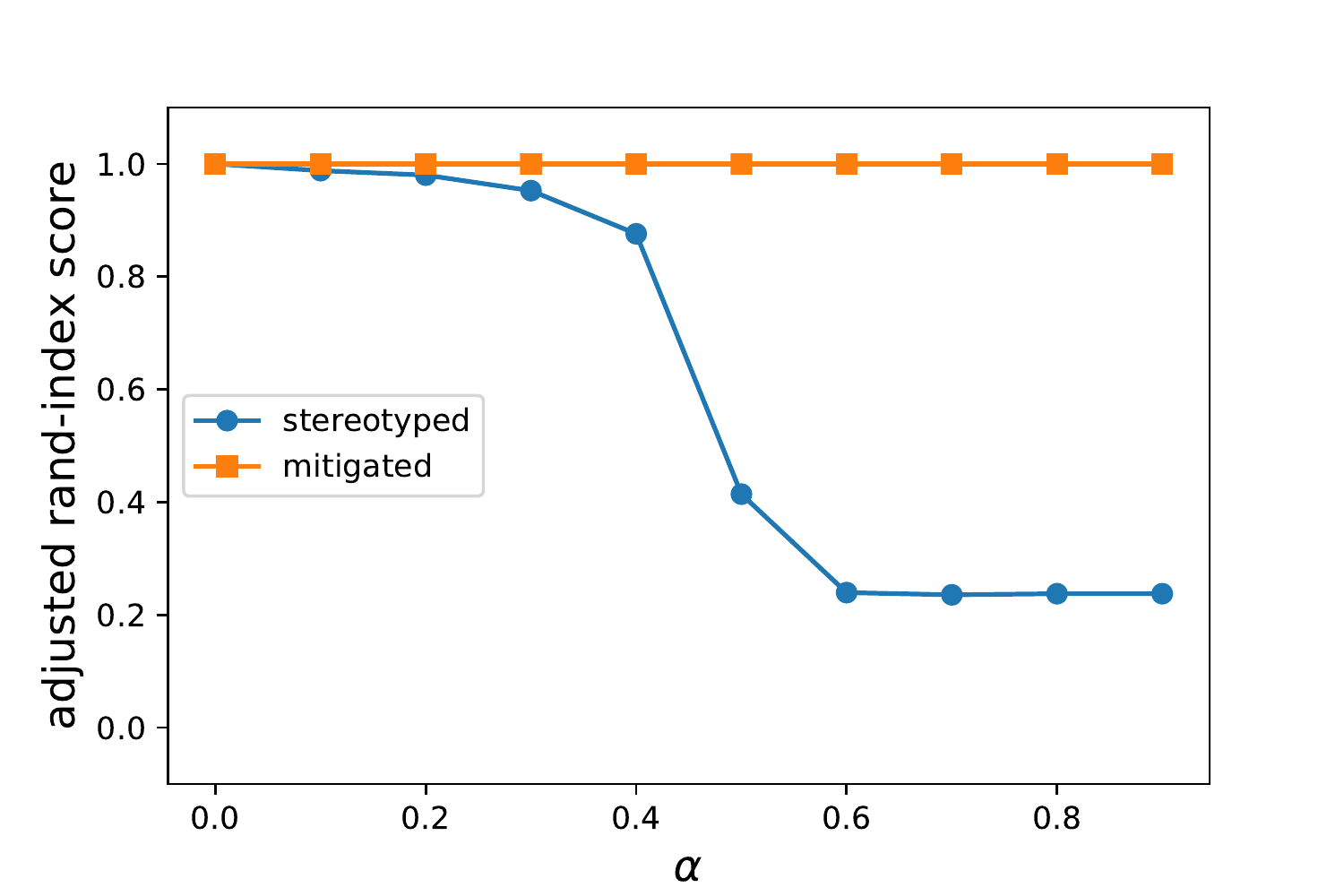} \label{fig:clustrand}}%
    ~~~~
    \subfloat[By boosting $\alpha$, clustering gets less balanced.]{\includegraphics[width=0.45\columnwidth]{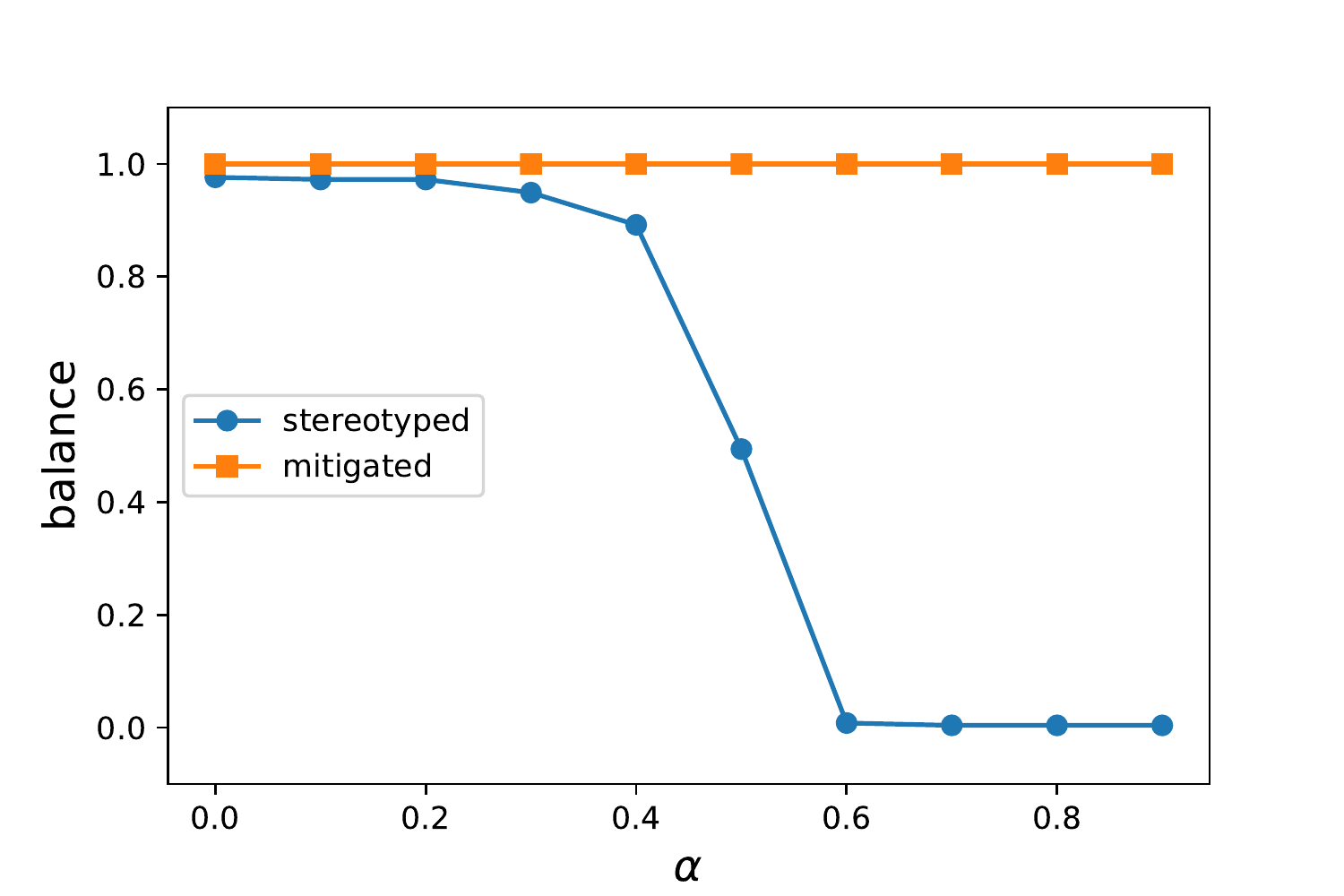} \label{fig:clustbalance}}\\
    \subfloat[Fair Clustering inflicts a higher cost than $K$-means]{\includegraphics[width=0.45\columnwidth]{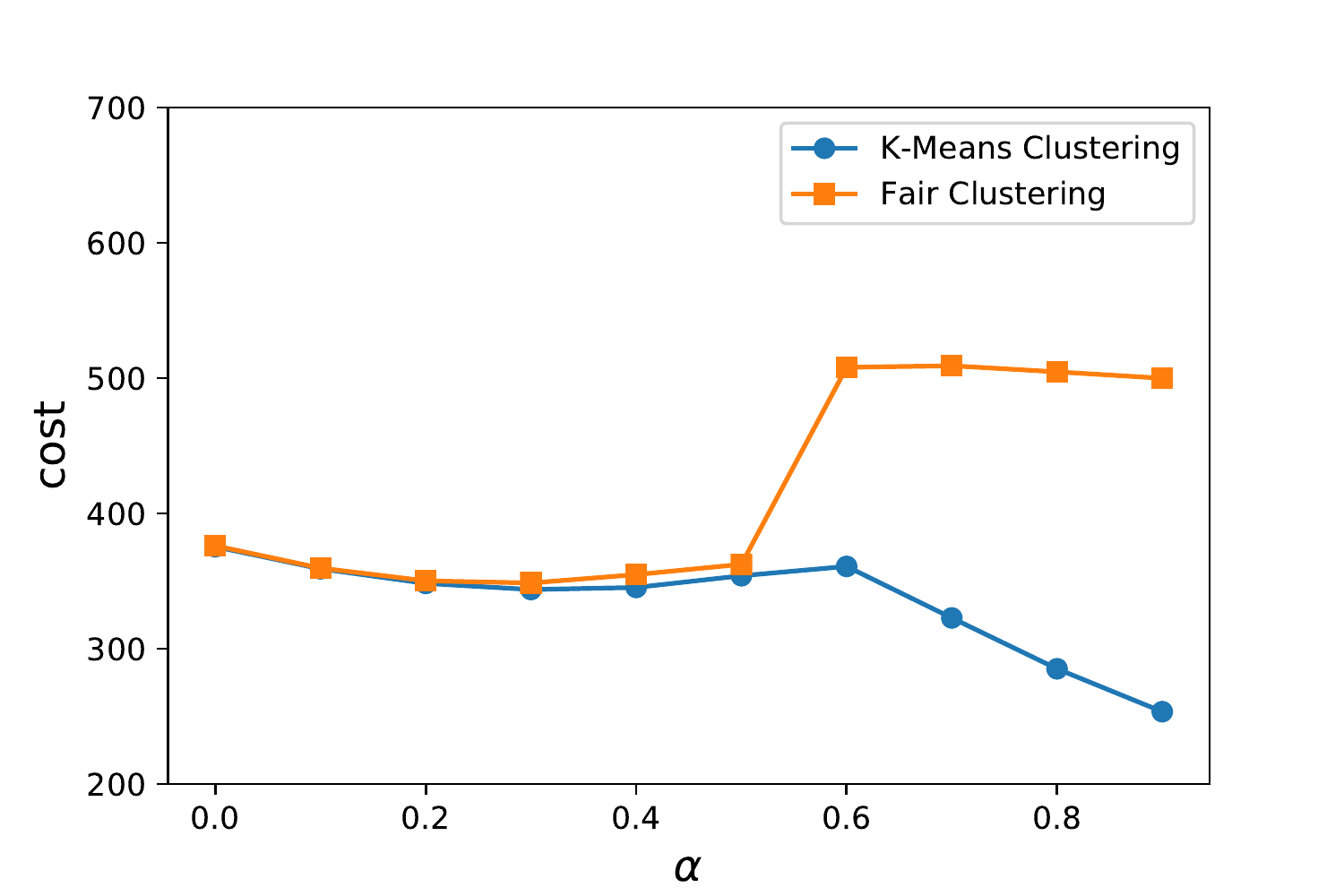} \label{fig:fairlet}}%
    ~~~~
    \subfloat[The two clustering impose the same cost after mitigation.]{\includegraphics[width=0.45\columnwidth]{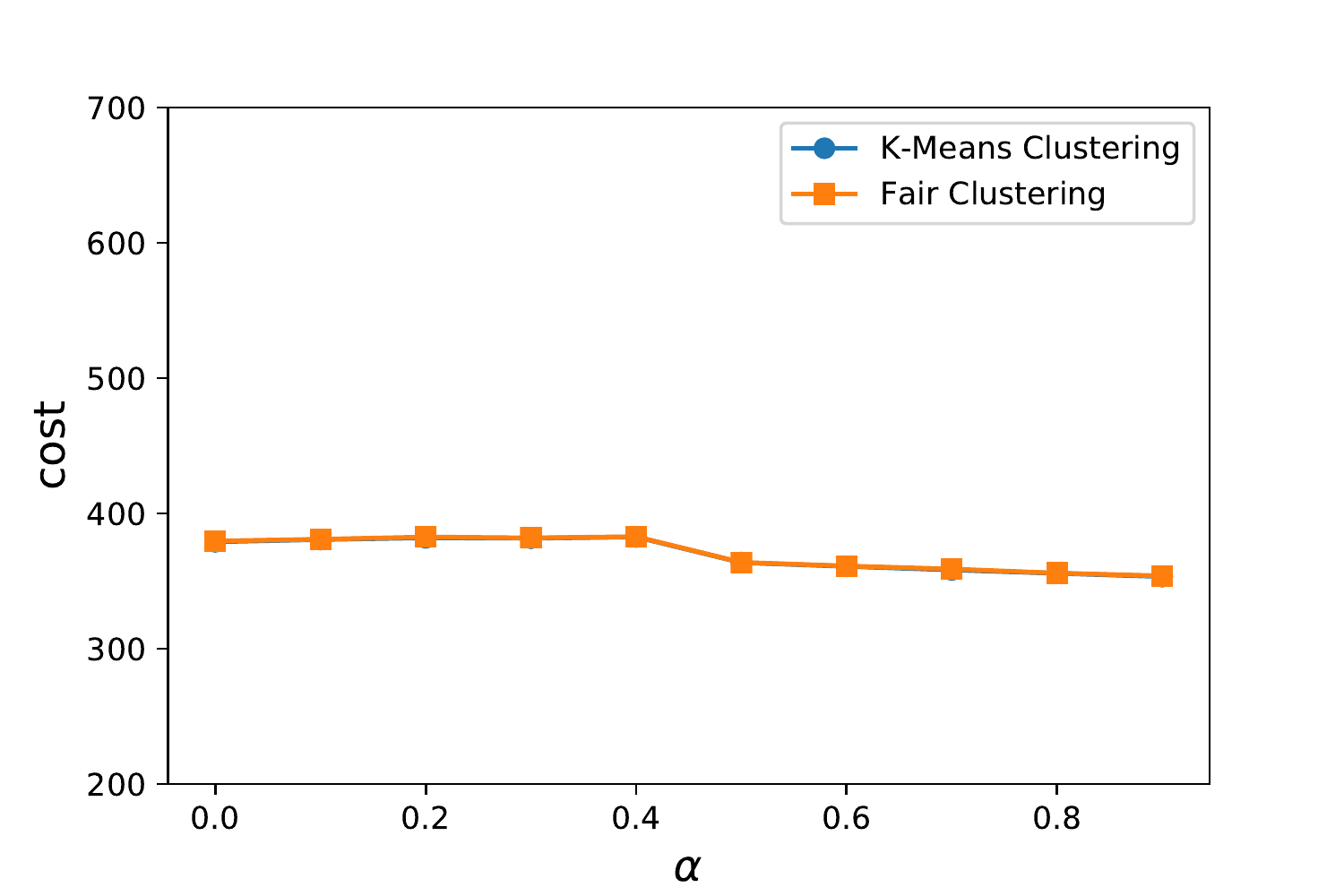} \label{fig:fairmitigate}}%
    \caption{Clustering in the presence of stereotyping.}%
    \label{fig:clustering-experiment}%
\end{figure}

In Figures \ref{fig:clustrand} and \ref{fig:clustbalance} we compare the results of $k$-means clusterings on stereotyped data and its mitigated representations for different values of $0 \leq \alpha \leq 1$.  We see a strong effect of the stereotyped representation for larger values of $\alpha$ and find that the mitigation strategy removes that effect.  This holds under both the rand-index score \cite{Hubert1985} and the balance notion proposed by \citep{chierichetti2017fair}.
In addition, although the solution proposed by Chierichetti et al. \citep{chierichetti2017fair} achieves a fair clustering, it would increase the cost of clustering in presence of stereotyping. This higher cost for fair clustering compared to $k$-means and the effectiveness of mitigation strategy are illustrated in Figures \ref{fig:fairlet} and \ref{fig:fairmitigate} respectively.
\vspace*{-0.2in}

%%% Local Variables:
%%% mode: latex
%%% TeX-master: "stereo"
%%% End:

\section{Discussion and Conclusion}
\label{sec:discussion}

In this paper, we formalized two notions of stereotyping and demonstrated how stereotyped representations lead to skewed outcomes when part of a machine learning pipeline.  We also presented mitigation strategies for these stereotype definitions and demonstrated via experiments on synthetic data that these strategies could largely remove the stereotype effects added to the data.

There are many aspects of stereotyped data that our approach does not include.  One might assume that stereotyping (via exemplar or representativeness) might act more weakly on some individuals (e.g., celebrities) than others.  Extensions to this framework that allow the stereotyping effect to act on a subset of the unprivileged group or that allow variations in $\alpha$ based, e.g., on the distance from the exemplar would be interesting to consider.  We consider only a single exemplar, rather than many.  Additionally, while there has been work validating the representativeness model of stereotyping in human subjects \cite{bordalo2016stereotypes}, the specific geometric model of stereotyping via exemplars that we consider here has not been similarly validated.  There are also other cognitive models of stereotyping that we did not consider. % Finally, there are many other forms of representational harm not considered here, including many described in Crawford's talk \cite{crawford}. 

%%% Local Variables:
%%% mode: latex
%%% TeX-master: "stereo"
%%% End:

\small
\bibliography{refs}

\begin{thebibliography}{10}

\bibitem{arrow1973theory}
K.~Arrow.
\newblock The theory of discrimination.
\newblock {\em Discrimination in labor markets}, 3(10):3--33, 1973.

\bibitem{becker1957economics}
G.~S. Becker.
\newblock The economics of discrimination, 1957.

\bibitem{Bera19}
S.~K. Bera, D.~Chakrabarty, and M.~Negahbani.
\newblock Fair algorithms for clustering.
\newblock {\em arXiv preprint arXiv:1901.02393}, 2019.

\bibitem{Bolukbasi:2016:MCP:3157382.3157584}
T.~Bolukbasi, K.-W. Chang, J.~Zou, V.~Saligrama, and A.~Kalai.
\newblock Man is to computer programmer as woman is to homemaker? debiasing
  word embeddings.
\newblock In {\em Proc. 30th NIPS}, NIPS'16, pages 4356--4364, 2016.

\bibitem{bordalo2016stereotypes}
P.~Bordalo, K.~Coffman, N.~Gennaioli, and A.~Shleifer.
\newblock Stereotypes.
\newblock {\em The Quarterly Journal of Economics}, 131(4):1753--1794, 2016.

\bibitem{brunet2018}
M.-E. Brunet, C.~Alkalay-Houlihan, A.~Anderson, and R.~Zemel.
\newblock Understanding the {Origins} of {Bias} in {Word} {Embeddings}.
\newblock {\em arXiv:1810.03611 [cs, stat]}, Oct. 2018.
\newblock arXiv: 1810.03611.

\bibitem{Calders10NaiveBayes}
T.~Calders and S.~Verwer.
\newblock Three naive bayes approaches for discrimination-free classification.
\newblock {\em Data Mining journal; special issue with selected papers from
  ECML/PKDD}, 2010.

\bibitem{caliskan_semantics_2017}
A.~Caliskan, J.~J. Bryson, and A.~Narayanan.
\newblock Semantics derived automatically from language corpora contain
  human-like biases.
\newblock {\em Science}, 356(6334):183--186, Apr. 2017.

\bibitem{cantor1979prototypes}
N.~Cantor and W.~Mischel.
\newblock Prototypes in person perception.
\newblock In {\em Advances in experimental social psychology}, volume~12, pages
  3--52. Elsevier, 1979.

\bibitem{cardwell1999}
M.~Cardwell.
\newblock {\em The Dictionary of Psychology}.
\newblock Taylor \& Francis, 1999.

\bibitem{chatterjee2009sensitivity}
S.~Chatterjee and A.~S. Hadi.
\newblock {\em Sensitivity analysis in linear regression}, volume 327.
\newblock John Wiley \& Sons, 2009.

\bibitem{chierichetti2017fair}
F.~Chierichetti, R.~Kumar, S.~Lattanzi, and S.~Vassilvitskii.
\newblock Fair clustering through fairlets.
\newblock In {\em Proc. NIPS}, pages 5029--5037, 2017.

\bibitem{crawford}
K.~Crawford.
\newblock The trouble with bias.
\newblock \url{https://www.youtube.com/watch?v=fMym_BKWQzk}, December 2017.

\bibitem{de-arteaga_bias_2019}
M.~De-Arteaga, A.~Romanov, H.~Wallach, J.~Chayes, C.~Borgs, A.~Chouldechova,
  S.~Geyik, K.~Kenthapadi, and A.~T. Kalai.
\newblock Bias in {Bios}: {A} {Case} {Study} of {Semantic} {Representation}
  {Bias} in a {High}-{Stakes} {Setting}.
\newblock In {\em Proceedings of the {Conference} on {Fairness},
  {Accountability}, and {Transparency}}, {FAT}* '19, pages 120--128, New York,
  NY, USA, 2019. ACM.

\bibitem{diaz2007exact}
J.~A. D{\i}az-Garc{\i}a, G.~Gonz{\'a}lez-Far{\i}as, and V.~M. Alvarado-Castro.
\newblock Exact distributions for sensitivity analysis in linear regression.
\newblock {\em Applied Mathematical Sciences}, 1(22):1083--1100, 2007.

\bibitem{edwards2015}
H.~Edwards and A.~Storkey.
\newblock Censoring {Representations} with an {Adversary}.
\newblock {\em arXiv:1511.05897 [cs, stat]}, Nov. 2015.
\newblock arXiv: 1511.05897.

\bibitem{Feldman2015DisparateImpact}
M.~Feldman, S.~A. Friedler, J.~Moeller, C.~Scheidegger, and
  S.~Venkatasubramanian.
\newblock Certifying and removing disparate impact.
\newblock {\em Proc. of KDD}, pages 259--268, 2015.

\bibitem{friedler2016possibility}
S.~A. Friedler, C.~Scheidegger, and S.~Venkatasubramanian.
\newblock On the (im) possibility of fairness.
\newblock {\em arXiv preprint arXiv:1609.07236}, 2016.

\bibitem{hardt2016equality}
M.~Hardt, E.~Price, N.~Srebro, et~al.
\newblock Equality of opportunity in supervised learning.
\newblock In {\em Proc. NIPS}, pages 3315--3323, 2016.

\bibitem{hilton1996stereotypes}
J.~L. Hilton and W.~Von~Hippel.
\newblock Stereotypes.
\newblock {\em Annual review of psychology}, 47(1):237--271, 1996.

\bibitem{Hubert1985}
L.~Hubert and P.~Arabie.
\newblock Comparing partitions.
\newblock {\em Journal of Classification}, 2(1):193--218, Dec 1985.

\bibitem{kahneman1972subjective}
D.~Kahneman and A.~Tversky.
\newblock Subjective probability: A judgment of representativeness.
\newblock {\em Cognitive psychology}, 3(3):430--454, 1972.

\bibitem{kahneman1973psychology}
D.~Kahneman and A.~Tversky.
\newblock On the psychology of prediction.
\newblock {\em Psychological review}, 80(4):237, 1973.

\bibitem{Kamishima12Fairness}
T.~Kamishima, S.~Akaho, H.~Asoh, and J.~Sakuma.
\newblock Fairness-aware classifier with prejudice remover regularizer.
\newblock {\em Machine Learning and Knowledge Discovery in Databases}, pages
  35--50, 2012.

\bibitem{katz1933verbal}
D.~Katz and K.~W. Braly.
\newblock Verbal stereotypes and racial prejudice.
\newblock {\em Journal of abnormal and social psychology}, 28:280--290, 1933.

\bibitem{Madras2018}
D.~Madras, E.~Creager, T.~Pitassi, and R.~Zemel.
\newblock Learning adversarially fair and transferable representations.
\newblock Technical report, arXiv preprint arXiv:1802.06309, 2018.

\bibitem{manis1988stereotypes}
M.~Manis, T.~E. Nelson, and J.~Shedler.
\newblock Stereotypes and social judgment: Extremity, assimilation, and
  contrast.
\newblock {\em Journal of Personality and Social Psychology}, 55(1):28, 1988.

\bibitem{olfat_convex_2018}
M.~Olfat and A.~Aswani.
\newblock Convex {Formulations} for {Fair} {Principal} {Component} {Analysis}.
\newblock {\em arXiv:1802.03765 [cs, math, stat]}, Feb. 2018.
\newblock arXiv: 1802.03765.

\bibitem{Romei13Multidisciplinary}
A.~Romei and S.~Ruggieri.
\newblock A multidisciplinary survey on discrimination analysis.
\newblock {\em The Knowledge Engineering Review}, pages 1--57, April 3 2013.

\bibitem{samadi_price_2018}
S.~Samadi, U.~Tantipongpipat, J.~H. Morgenstern, M.~Singh, and S.~Vempala.
\newblock The {Price} of {Fair} {PCA}: {One} {Extra} dimension.
\newblock In {\em Proc. NeurIPS}, pages 10999--11010, 2018.

\bibitem{Schmidt18}
M.~Schmidt, C.~Schwiegelshohn, and C.~Sohler.
\newblock Fair coresets and streaming algorithms for fair k-means clustering.
\newblock {\em arXiv preprint arXiv:1812.10854}, 2018.

\bibitem{smith1992exemplar}
E.~R. Smith and M.~A. Zarate.
\newblock Exemplar-based model of social judgment.
\newblock {\em Psychological review}, 99(1):3, 1992.

\bibitem{speer}
R.~Speer.
\newblock Conceptnet numberbatch 17.04: better, less-stereotyped word vectors.
\newblock \url{http://bit.ly/speer-concept}, Apr 2017.

\bibitem{statt2002}
D.~Statt.
\newblock {\em The concise dictionary of psychology}.
\newblock Routledge, 2002.

\bibitem{sweeney2013discrimination}
L.~Sweeney.
\newblock Discrimination in online ad delivery.
\newblock {\em Queue}, 11(3):10, 2013.

\bibitem{tversky1974judgment}
A.~Tversky and D.~Kahneman.
\newblock Judgment under uncertainty: Heuristics and biases.
\newblock {\em Science}, 185(4157):1124--1131, 1974.

\bibitem{zliobaite2015survey}
I.~\v{Z}liobait\.{e}.
\newblock A survey on measuring indirect discrimination in machine learning.
\newblock {\em arXiv preprint arXiv:1511.00148}, 2015.

\bibitem{woodbury}
{Wikipedia contributors}.
\newblock Woodbury matrix identity.
\newblock
  \url{https://en.wikipedia.org/w/index.php?title=Woodbury_matrix_identity&oldid=856202501},
  2018.

\bibitem{zafar2017mistreatment}
M.~B. Zafar, I.~Valera, M.~G. Rodriguez, and K.~P. Gummadi.
\newblock Fairness beyond disparate treatment \& disparate impact: Learning
  classification without disparate mistreatment.
\newblock {\em Proc. of WWW}, pages 1171--1180, 2017.

\bibitem{icml2013_zemel13}
R.~Zemel, Y.~Wu, K.~Swersky, T.~Pitassi, and C.~Dwork.
\newblock Learning fair representations.
\newblock In {\em Proc. of Intl. Conf. on Machine Learning}, pages 325--333,
  2013.

\end{thebibliography}
\bibliographystyle{abbrv}
\end{document}